\documentclass[preprint,3p,12pt]{elsarticle}

\usepackage{amsmath,amssymb,amsfonts}
\usepackage{graphicx, enumitem}
\usepackage{textcomp}
\usepackage[ruled,vlined,linesnumbered]{algorithm2e}
\usepackage[table]{xcolor}
\usepackage{tikz}
\usepackage{bm}
\usepackage{booktabs}
\usepackage{multirow}
\usepackage{url}
\usepackage[left, modulo]{lineno}
\usepackage{subfig}

\SetKwInOut{Parameter}{Params}

\newcolumntype{L}[1]{>{\raggedright\let\newline\\\arraybackslash\hspace{0pt}}m{#1}}
\newcolumntype{C}[1]{>{\centering\let\newline\\\arraybackslash\hspace{0pt}}m{#1}}
\newcolumntype{R}[1]{>{\raggedleft\let\newline\\\arraybackslash\hspace{0pt}}m{#1}}

\journal{Information Sciences}

\begin{document}
	
\begin{frontmatter}
	

\title{AT-MFCGA: An Adaptive Transfer-guided Multifactorial Cellular Genetic Algorithm for Evolutionary Multitasking}

\author[mymainaddress]{Eneko~Osaba\corref{mycorrespondingauthor}}
\ead{eneko.osaba@tecnalia.com}
\author[mymainaddress,mythirdaddress]{Javier Del Ser}
\author[mymainaddress]{Aritz~D.~Martinez}
\author[mymainaddress]{\\Jesus~L.~Lobo}
\author[myfourthaddress]{Francisco~Herrera}

\cortext[mycorrespondingauthor]{Corresponding author: TECNALIA. Parque Tecnologico, Ed. 700, 48160 Derio, Bizkaia, Spain. Telephone: (+34) 946 430 850. Fax: (+34) 901 760 009}
\address[mymainaddress]{TECNALIA, Basque Research and Technology Alliance (BRTA), \\48160 Derio, Bizkaia, Spain}
\address[mythirdaddress]{Basque Center for Applied Mathematics (BCAM), 48009 Bilbao, Spain}
\address[myfourthaddress]{DaSCI Andalusian Institute of Data Science and Computational Intelligence, \\University of Granada, Spain}

\begin{abstract}
Transfer Optimization is an incipient research area dedicated to solving multiple optimization tasks simultaneously. Among the different approaches that can address this problem effectively, Evolutionary Multitasking resorts to concepts from Evolutionary Computation to solve multiple problems within a single search process. In this paper we introduce a novel adaptive metaheuristic algorithm to deal with Evolutionary Multitasking environments coined as Adaptive Transfer-guided Multifactorial Cellular Genetic Algorithm (AT-MFCGA). AT-MFCGA relies on cellular automata to implement mechanisms in order to exchange knowledge among the optimization problems under consideration. Furthermore, our approach is able to explain by itself the synergies among tasks that were encountered and exploited during the search, which helps us to understand interactions between related optimization tasks. A comprehensive experimental setup is designed to assess and compare the performance of AT-MFCGA to that of other renowned evolutionary multitasking alternatives (MFEA and MFEA-II). Experiments comprise 11 multitasking scenarios composed of 20 instances of 4 combinatorial optimization problems, yielding the largest discrete multitasking environment solved to date. Results are conclusive in regard to the superior quality of solutions provided by AT-MFCGA with respect to the rest of the methods, which are complemented by a quantitative examination of the genetic transferability among tasks throughout the search process.
\end{abstract}

	
\end{frontmatter}

\section{Introduction} \label{sec:intro}

Inspired by the same rationale that underlies Transfer Learning and Multitask Learning, Transfer Optimization (TO, \cite{gupta2017insights}) is currently gaining momentum within the research community \cite{ong2016evolutionary}. The main motivation behind this paradigm is that real-world optimization problems hardly ever occur in isolation. Consequently, TO aims to exploit the knowledge learned throughout the optimization of one problem (\emph{task}) when solving another related or unrelated problem or task. Since it can be regarded as a relatively new research stream, the community has only recently started to consider this transferability of knowledge among problems to be a research priority. Besides the co-occurrence of optimization problems in practice, other fundamental reasons for this increasing interest in TO include the growing scales and level complexity of such optimization tasks, which has led to the need to use previously obtained knowledge.

Three different categories of TO can be distinguished in the literature: \textit{sequential transfer}, \textit{multitasking} and \textit{multiform optimization}. The first one is related to tasks that are solved sequentially. The knowledge collected when facing preceding tasks is harnessed as external information to be used for the optimization of new instances/problems. The second category is referred to as \textit{multitasking}, which addresses different yet equally important tasks in a simultaneous fashion. In this category, the dynamic exploitation of synergies among problems is a crucial issue. Lastly, \textit{multiform optimization} regards the simultaneous tackling of a single task under different alternative problem formulations. Since the inception of the field, \textit{multitasking} stands out as arguably the most prolific research strand. 

Within the taxonomy described above, this paper focuses on Evolutionary Multitasking (EM, \cite{ong2016towards}), which relies on operators and search procedures from Evolutionary Computation \cite{del2019bio} to realize efficient multitasking methods. Many efforts have been reported in the recent literature trying to solve a variety of discrete, continuous, single-objective and multi-objective optimization problems through the perspective of EM \cite{wang2019evolutionary,gong2019evolutionary}. From the algorithmic point of view, most of the current research related to EM is materialized by embracing a Multifactorial Optimization (MFO) strategy. In MFO a unique factor is established for each solution in order to determine its specialization with respect to the set of problems being solved, ultimately driving the search of population-based methods. The combination of MFO and EM concepts has given rise to the renowned Multifactorial Evolutionary Algorithm (MFEA, \cite{gupta2015multifactorial}), which is undoubtedly at the forefront of the techniques introduced so far in the TO field.

Although TO is a relatively young research field, there is consensus in the community regarding the capital relevance of the correlation among problems, particularly in multitasking \cite{gupta2015multifactorial}. The exploitation of these interrelationships is paramount to positively capitalize the transfer of valuable knowledge during the search \cite{zhou2018study}. Several influential studies have been published recently that analyze this issue. Some of these studies introduce alternatives that properly quantify the similarities among optimization tasks and problems \cite{gupta2016landscape}. Nevertheless, in many practical circumstances and scenarios all problems cannot be guaranteed to be strictly related to each other. When no such synergies exist, sharing genetic material between unrelated tasks usually leads to performance downturns (\textit{negative transfer}, \cite{bali2019multifactorial}). This phenomenon has been reported in recent studies as the main pitfall of multitasking approaches, as it is a common challenge in the modeling of new solving schemes. In this context, the Multifactorial Evolutionary Algorithm II (MFEA-II, \cite{bali2019multifactorial}) incorporates adaptive mechanisms for dynamically learning how much knowledge should be transferred across different problems. As occurred with MFEA in the past, MFEA-II has emerged as a baseline of new adaptive TO schemes, introducing new algorithmic ingredients that make its search resilient against negative information transfer.

Keeping all this in mind, the principal objective of this paper is to propose a new MFO approach called \textit{Adaptive Transfer-guided Multifactorial Cellular Genetic Algorithm} (AT-MFCGA). AT-MFCGA relies on the foundations of cellular automata and Cellular Genetic Algorithms (cGAs \cite{manderick1989fine}) for controlling the mating process among individuals. This method builds on preliminary research in \cite{osaba2020multifactorial} and improves it significantly by addressing the need to dynamically identify and exploit the synergies between tasks that arise during the search process. Specifically, the two key elements of this paper with respect to the state of the art and our preliminary findings in \cite{osaba2020multifactorial} can be summarized as follows:
\begin{itemize}[leftmargin=*]
	\item AT-MFCGA is able to assess the performance of mating operations conducted among individuals specialized in different tasks, quantifying both negative and positive transfers along the search. Based on this information, the cellular grid in which the population is organized is rearranged to promote crossovers among individuals specialized in tasks expected to yield positive knowledge transfer. In addition, we introduce a mechanism to also adapt local search operators based on this information.
	
	\item The search strategy of AT-MFCGA allows for a quantitative examination of synergies that arise among tasks. This feature provides a novel explainability interface for the user to better understand the interactions between problems. We take advantage of this characteristic by further examining the genetic transferability among the problem instances used in our experiments. This analysis is a valuable addition to the state of the art \cite{zhou2018study,gupta2016landscape}, and provides useful insights for further research. In addition, we provide an additional visualization of the solver's performance, visually depicting the influence of the genetic complementarities on the rebuilt cellular grid.
\end{itemize}

An extensive set of experiments is reported using instances of 4 combinatorial optimization problems, namely, Traveling Salesman Problem (TSP, \cite{lawler1985traveling}), Capacitated Vehicle Routing Problem (CVRP, \cite{toth2002vehicle}), Quadratic Assignment Problem (QAP, \cite{lawler1963quadratic}) and Linear Ordering Problem (LOP, \cite{bertsimas1997introduction}). The experimental setup comprises 11 test cases, using 20 different instances (5 for each combinatorial problem). To the best of our knowledge, the experimentation presented and discussed in this paper is one of the most extensive and detailed in discrete MFO. Moreover, we compare the performance of AT-MFCGA to that of MFEA \cite{gupta2015multifactorial} and MFEA-II \cite{bali2019multifactorial}. The obtained results conclude that AT-MFCGA outperforms the rest of algorithms in the benchmark with statistical significance, eliciting visual insights about the positive and negative genetic transfers held during the search. 

In summary, the main contribution of this paper is threefold: i) the design and implementation of a novel Adaptive Transfer-guided Multifactorial Cellular Genetic Algorithm for multitask optimization problems; ii) a deep examination of the genetic knowledge transfer among the instances of optimization problems of a diverse nature (TSP, VRP, QAP and LOP); and iii) an extensive experimentation composed of 20 instances belonging to 4 different combinatorial optimization problems.

The rest of the paper is organized as follows: Section \ref{sec:back} presents a brief overview of the background related to EM, cGAs, MFEA and recent solvers dealing with the negative transfer phenomenon. Section \ref{sec:MFCGA} details the main characteristics of AT-MFCGA. The description of the designed experimental setup is given in Section \ref{sec:exp}. Experimental results are presented and discussed in \ref{sec:exp_res}, along with a detailed analysis of the genetic transfer between tasks. Section \ref{sec:exp_rebuilding} examines the grid rebuilding mechanism of AT-MFCGA. Finally, Section \ref{sec:conc} concludes the paper by drawing conclusions and outlining several future lines of research derived from this work.

\section{Background} \label{sec:back}

This section offers a brief background of the four main aspects addressed in this study: EM and MFO (Section \ref{sec:back_EM}), MFEA (Section \ref{sec:MFEA}), MFEA-II and adaptive EM solvers (Section \ref{sec:MFEA-II}), and cGAs (Section \ref{sec:back_CGA}).

\subsection{Evolutionary Multitasking and Multifactorial Optimization} \label{sec:back_EM}

As previously mentioned, multitasking optimization is dedicated to simultaneously tackling several problems or tasks. Thus, this branch of TO is characterized by omni-directional knowledge sharing among different problems, potentially reaching a synergistic completion among the tasks being solved \cite{gupta2017insights}. EM has arisen as an efficient approach for dealing with these simultaneous optimization environments. 

Two main features encouraged researchers to formulate the EM paradigm. First, the inherent parallelism enabled by a population of individuals that are evolved together is well suited for simultaneously dealing with concurrent optimization problems. Some recently published studies have reported the benefits of this simultaneous treatment for dynamically unveiling latent relationships among problems \cite{ong2016evolutionary}. The second crucial feature of EM is the uninterrupted sharing of genetic material during the evolutionary search, which permits all tasks to benefit from one another \cite{gupta2015multifactorial}. As such, there are several methods used to deal with multitasking scenarios through the perspective of EM, the two most used approaches being: parallel search processes (typical in Multipopulation-based Multitasking) or the execution of a single search procedure (as in MFO).

A common point of agreement in the related literature is that until \cite{da2017evolutionary}, the EM paradigm was only materialized through the perspective of MFO. Several approaches have since then embraced this concept, encompassing hybrid solvers \cite{xiao2019multifactorial}, modern metaphors \cite{zheng2016multifactorial}, or multipopulation schemes \cite{li2020multifactorial}. Furthermore, other alternatives to MFO have also been proposed in the form of new algorithmic schemes, such as coevolutionary multitasking or multitasking multi-swarm optimization \cite{song2019multitasking}. Despite this recent upsurge of new MFO and EM frameworks, MFEA is still the spearhead of the field \cite{gupta2015multifactorial}.  

Mathematically, MFO is described as an EM environment comprising $K$ optimization tasks $\{T_k\}_{k=1}^K$ to be solved in a simultaneous manner. This scenario is made up by as many search spaces as problems being considered, each one corresponding to a single task $T_k$. Moreover, the objective function for the $k$-th task is represented as $f_k : \Omega_k \rightarrow \mathbb{R}$, where $\Omega_k$ is the search space of $T_k$. Let us assume that all tasks are minimization problems. Thus, the main goal of MFO is to find a set of solutions $\{\mathbf{x}_1,\dots,\mathbf{x}_K\}$ such that $\mathbf{x}_k = \arg \min_{\mathbf{x}\in\Omega_k} f_k(\mathbf{x})$. Instead of facing all these tasks via independent search processes, MFO solvers work with a population $\mathcal{P}$ of candidate solutions, with each $\mathbf{x}_p\in\mathcal{P}$ belonging to a unified search space $\Omega_U$. This way, each independent search space $\Omega_k$ belonging to a task $k$ can be translated to $\Omega_U$ using an encoding/decoding function $\xi_k: \Omega_k\mapsto \Omega_U$. As a consequence, each individual $\mathbf{x}_p\in\mathcal{P}$ must be encoded as $\mathbf{x}_{p,k}=\xi_k^{-1}(\mathbf{x}_p)$ in order to properly represent a task-specific solution $\mathbf{x}_{p,k}$ for each of the $K$ problems. 

It is interesting here to delve further into the formulation of the common search space $\Omega_U$, which is arguably one of the key aspects of EM. This $\Omega_U$ should be consistent with the level overlapping between tasks being addressed. In this regard, and based on the research conducted by Ong and Gupta in \cite{ong2016evolutionary}, we can assess the overlap ($\chi$) of two tasks based on the number of variables in the task-specific solution space which have the same phenotypic meaning, i.e., $\chi$ = $|x_{overlap}|$. This way, we can identify three superposition categories of overlap in the phenotype space: i) complete overlap ($x_1 \ x_{overlap}$ = $x_2 \ x_{overlap}$ = $\emptyset$ ), when problems to solve differ only in their task-specific auxiliary variables; ii) partial overlap ($x_1 \ x_{overlap}$ $\neq$ $\emptyset$ and/or $x_2 \ x_{overlap}$ $\neq$ $\emptyset$ and $\chi$ $\geqslant$ 1), for problems in which the distribution of variables is similar, or when tasks have some recurrent variables in common; and iii) no overlap ($x_{overlap}$ = $\emptyset$), when tasks do not have any structural characteristics in common.

In addition to the above notation, MFO algorithms rely on four different specific concepts, associated to each solution $\mathbf{x}_p^\prime\in\Omega^\prime$ of the $P$ population:
\begin{itemize}[leftmargin=*]
\item \textit{Factorial Cost}: the factorial cost $\Psi_k^p$ of an individual $\mathbf{x}_p^\prime$ is the fitness value for a specific task $T_k$. 

\item \textit{Factorial Rank}: the factorial rank $r_k^p$ of a population member $\mathbf{x}_p$ in a task $T_k$ is the rank of this member within the whole population, sorted in ascending order of $\Psi_k^p$. 

\item \textit{Scalar Fitness}: the scalar fitness $\varphi^p$ of an individual $\mathbf{x}_p^\prime$ is computed using the best $r_k^p$ over all the tasks, i.e., \smash{$\varphi^p = 1/ \left(\min_{k \in \{1...K\}}r_k^p\right)$}. 

\item \textit{Skill Factor}: the skill factor $\tau^p$ is the index of the task in which $\mathbf{x}_p^\prime$ performs best, namely, $\tau^p = \arg \min_{k\in\{1,\ldots,K\}} r_k^p$. 
\end{itemize}

These four definitions are the ones on which all MFO techniques rely. Going further, these concepts are employed with different purposes, such as i) assigning tasks to individuals ii) deciding how population members interact, iii) sorting and classifying the complete population and iv) deciding which solutions survive the iterations. As such, these definitions have led to the design and implementation of different efficient solving approaches. 

In addition, it is also worth-mentioning that two knowledge sharing patterns can be found in EM methods: \textit{implicit transfer} and \textit{explicit transfer}. On the one hand, in \textit{implicit transfer}, the sharing of genetic material is capitalized through search operators as crossover functions. On the other hand, \textit{explicit knowledge} sharing is principally materialized by migrating complete solutions among populations, namely from one task to another. Arguably, one of the most frequently occurring patterns is the first one, implicit transfer, which is the one used throughout this paper.

\subsection{Multifactorial Evolutionary Algorithm}\label{sec:MFEA}

MFEA,  a recently introduced MFO solver based on bio-cultural schemes of multifactorial inheritance, is grounded on the previously described concepts \cite{gupta2015multifactorial}. In Algorithm~\ref{alg:classicMFEA} we depict the main workflow of MFEA. In order not to dwell extensively on algorithmic aspects, we refer interested readers to the detailed description provided in \cite{gupta2015multifactorial}. Briefly explained, the search process of MFEA relies on four key concepts:
\begin{itemize}[leftmargin=*]
\item \textit{Unified solution representation}: one of the most crucial aspects when developing a MFEA is the representation strategy used to encode an individual $\mathbf{x}_i^\prime$, which yields the unified search space $\Omega^\prime$. The experimental benchmark proposed in this study can exemplify how a representation strategy should be designed. Since four different permutation based discrete optimization problems (TSP, CVRP, QAP and LOP) are considered, a permutation encoding strategy can be chosen as the unified representation of $\mathbf{x}_i^\prime$. In this regard, we have faithfully followed the procedure described in \cite{yuan2016evolutionary} for the individual representation. Thus, if $K$ problems are to be faced simultaneously, and denoting the dimension of each task $T_k$ as $D_k$, an individual $\mathbf{x}_i^\prime$ is represented as a permutation of the integer set $\{1,2,\ldots, D_{max}\}$, where $D_{max}=\max_{k\in\{1,\ldots,K\}} D_k$, that is, the maximum dimension among all the considered $K$ tasks. Therefore, when $\mathbf{x}_i^\prime$ is to be measured in the task $T_k$ whose $D_k<D_{max}$, only those values lower than $D_k$ are considered for building the argument solution $\mathbf{x}_k$ of $f_k(\cdot)$. This reconstruction is carried out by maintaining the same order as in $\mathbf{x}_i^\prime$. Other unified encoding approaches utilized in the literature include, among others, random keys representation \cite{da2017evolutionary}.

\item \textit{Assortative Mating}: this characteristic establishes that individuals prioritize relationships with mates belonging to the same cultural background \cite{gupta2015multifactorial}. This way, genetic operators used for implementing a MFEA should encourage the cross between individuals with the same skill factor $\tau^i$. Further technical details on the implementation of this mechanism can be found in the aforementioned studies.

\item \textit{Selective evaluation}: this mechanism states that each generated offspring is assessed in just one task, instead of being measured in each task separately. A newly created individual is evaluated in task $T_{\tau_\ast^i}$, where $\tau_\ast^i$ is the skill factor of its parent. In case the offspring has more than one parent, $\tau_\ast^i$  is randomly selected among their skill factors. Furthermore, the factorial cost $\Psi_k^i$ is set to $\infty$ $\forall k\in\{1,\ldots,\tau_\ast^i-1,\tau_\ast^i+1,\ldots,K\}$. 

\item \textit{Scalar fitness based selection}: analogously to canonical Genetic Algorithms, this feature represents the survivor function of the MFEA. In this case, the selection is based on an elitist strategy, i.e. the best $P$ individuals in terms of scalar fitness are those selected for survival and are passed on to the next generation.
\end{itemize}
\begin{algorithm}[h!]
	\SetAlgoLined
	\DontPrintSemicolon
	Randomly draw a population of $|\mathcal{P}|=P$ individuals $\{\mathbf{x}_i\}_{i=1}^P$, with $\mathbf{x}_i\in\Omega_U$\;
	Evaluate each generated individual for the $K$ problems\;
	Calculate the skill factor $\tau_i$ of each $\mathbf{x}_i$\;
	\While{termination criterion not reached}{
		Set $\mathcal{Q}=\emptyset$\;
		\While{individuals still to be selected}{
			Randomly sample without replacement $\mathbf{x}_{i'},\mathbf{x}_{i''}\in\mathcal{P}$\;
			\uIf{$\tau_{i'}=\tau_{i''}$}{
				$[\mathbf{x}_{A},\mathbf{x}_{B}] = \mbox{IntrataskCX}(\mathbf{x}_{i'},\mathbf{x}_{i''})$\;
				Set $\tau_A$ and $\tau_B$ equal to $\tau_{i'}$\;
			}\uElseIf{$rand\leq \mbox{RMP}$}{
				$[\mathbf{x}_{A},\mathbf{x}_{B}] = \mbox{IntertaskCX}(\mathbf{x}_{i'},\mathbf{x}_{i''})$\;
				Set $\tau_A=rand(\tau_{i'},\tau_{i''})$ and $\tau_B=rand(\tau_{i'},\tau_{i''})$\;
			}\uElse{
				$\mathbf{x}_A = \mbox{mutation}(\mathbf{x}_{i'})$;  \smash{$\tau_A=\tau_{i'}$}\;
				$\mathbf{x}_B = \mbox{mutation}(\mathbf{x}_{i''})$; \smash{$\tau_B=\tau_{i''}$}\;
			}
			Evaluate $\mathbf{x}_A$ for task $\tau_A$, and $\mathbf{x}_B$ for task $\tau_B$\;
			$\mathcal{Q}$ = $\mathcal{Q}\cup\{\mathbf{x}_A,\mathbf{x}_B\}$\;
		}
		Select the best $P$ individuals in $\mathcal{P}\cup\mathcal{Q}$ as per their $\varphi_i$ \;
	}
	Return the best individual in $\mathcal{P}$ for each task $T_k$\;
	\caption{Pseudocode of MFEA}
	\label{alg:classicMFEA}
\end{algorithm}

Since it was first reported in 2015, MFEA has been the focus of vibrant research activity. To begin with, in \cite{yuan2016evolutionary} MFEA was adapted to different discrete problems, such as TSP, LOP and QAP. Similar research was introduced in \cite{zhou2016evolutionary}, where MFEA was applied to the Vehicle Routing Problem. Another discrete problem -- Clustered Minimum Routing Cost Problem -- was also addressed with MFEA by Trung et al. in \cite{trung2019multifactorial}. More recent is the study proposed in \cite{tam2020multifactorial}, in which a MFEA in the context of wireless sensor networks is developed, aiming at maximizing data aggregation tree lifetime. Other applications of MFEA can be found in \cite{wang2019evolutionary} for the composition of semantic web services, and in \cite{martinez2020simultaneously} to evolve deep reinforcement learning models. Furthermore, an enhanced variant of the MFEA is introduced in \cite{gong2019evolutionary}, endowing the basic version of the algorithm with a dynamic resource allocation strategy. A similar philosophy is followed in \cite{yu2019multifactorial}, describing an improved MFEA by incorporating opposition-based learning. In \cite{gupta2016multiobjective}, a multiobjective version of MFEA is introduced (MO-MFEA), assessing its efficiency over continuous benchmark functions, as well as a real-world manufacturing process design problems. Also interesting is the study proposed in \cite{yang2017two} in which authors introduce a two-stage assortative mating mechanism for improving the performance of the MO-MFEA. A novel multi-objective approach is also presented in \cite{wang2019multiobjective}, in this case in form of an adaptive multiobjective and multifactorial differential evolution algorithm. Moreover, an improved MO-MFEA is proposed in \cite{yao2020multiobjective}, based on the decomposition and dynamic resource allocation strategy. In such paper, nine benchmark test cases are selected for experimental studies, each one comprising two-task problems.

\subsection{From MFEA to MFEA-II: searching for adaptability} \label{sec:MFEA-II}

Notwithstanding the success of MFEA, EM and the wider field of TO, these areas are the target of controversial discussions questioning the efficiency of methods proposed to date. These criticisms accentuate the complexity of avoiding, identifying and/or reacting against negative transfers between tasks \cite{wang2019rigorous}. As pointed out in the previous section, it is well established that to obtain the best performance and real potential of TO algorithms, the exploitation of the synergies between the problems is of paramount importance. This is the main reason why the related community is striving to propose new methods to cope with this situation, favoring positive transfers and making optimization algorithms resilient to the existence of negative influences among tasks \cite{liang2020two}. In fact, this is the main goal of the AT-MFCGA solver proposed in this study, and also the objective of the evolution of MFEA, which has been coined MFEA-II \cite{bali2019multifactorial}.

Accordingly, the principal contribution provided by MFEA-II with respect to its predecessor is the inclusion of a transfer parameter matrix (RMP matrix), which takes on the responsibility of determining the extent of genetic transfer across individuals with different \textit{skill tasks}. This transfer parameter matrix is dynamically updated based on the information generated during the course of the multitasking search. This RMP matrix is managed and updated by an \textit{online RMP learning module}. An additional characteristic of MFEA-II is the inter-task crossover procedure, which is composed by parent-centric operators \cite{deb2002real}. In short, these operators are conceived to generate solutions close to their parents in the search space (in this case, the unified search space $\Omega_U$). Based on the insights introduced in Bali et al.'s pioneering study, the main modifications of MFEA-II are concentrated in the inter-task crossover steps (lines 11-16 in Algorithm \ref{alg:classicMFEA}), which are replaced by those represented in Algorithm \ref{alg:MFEAIImodifiedsteps}. 
\begin{algorithm}[h!]
	\SetAlgoLined
	\DontPrintSemicolon
	\uIf{$\tau_{i'}\neq\tau_{i''}$}{
		\uIf{$rand\leq \mbox{RMP}_{\tau_{i'},\tau_{i''}}$}{
			$[\mathbf{x}_A,\mathbf{x}_B] = \mbox{IntertaskParentCentricCX}(\mathbf{x}_{i'},\mathbf{x}_{i''})$\;
			Update $\mathbf{x}_A = \mbox{mutation}(\mathbf{x}_A)$\;
			Update $\mathbf{x}_B = \mbox{mutation}(\mathbf{x}_B)$\;
			$\tau_A=rand(\tau_{i'},\tau_{i''})$\;
			$\tau_B=rand(\tau_{i'},\tau_{i''})$\;
		}\uElse{
			Randomly select $\mathbf{x}_{i1}\in\mathcal{P}$ with $\tau_{i1}=\tau_{i'}$\;
			$\mathbf{x}_A = \mbox{IntrataskParentCentricCX}(\mathbf{x}_{i'},\mathbf{x}_{i1})$\;
			Update $\mathbf{x}_A = \mbox{mutation}(\mathbf{x}_A)$\;
			$\tau_A=\tau_{i'}$\;
			Randomly select $\mathbf{x}_{i2}\in\mathcal{P}$ with $\tau_{i2}=\tau_{i''}$\;
			$\mathbf{x}_B = \mbox{IntertaskParentCentricCX}(\mathbf{x}_{i''},\mathbf{x}_{i2})$\;
			Update $\mathbf{x}_B = \mbox{mutation}(\mathbf{x}_B)$\;
			$\tau_B=\tau_{i''}$\;
		}
	}

	\caption{Inter-task crossover procedure of MFEA-II}
	\label{alg:MFEAIImodifiedsteps}
\end{algorithm}

Besides MFEA-II, a growing corpus of literature has recently been noted around dynamically and efficiently dealing with negative knowledge transfer in multitasking environments. That which is most directly related to MFEA-II is its multi-objective variant (MO-MFEA-II \cite{bali2020cognizant}), which follows the same concepts as its single-objective counterpart. The research proposed in \cite{yi2020multifactorial} follows this same line of thought, proposing a multi-objective \textit{novel interval} MFEA that embraces the $RMP$ online updating strategy of MFEA-II. A similar philosophy is followed in \cite{tang2019adaptive}, in which an adapted multifactorial particle swarm optimization method is proposed based on an inter-task learning parameter being updated during the search. Another solver introducing mechanisms to avoid negative transfers is the one presented in \cite{yin2019multifactorial}. In this case, authors implement the classical MFEA using a new inter-task knowledge transfer strategy. This new strategy is based on search direction instead of individuals, generating offspring as per the sum of an elite solution of one task and a difference vector from another task. In \cite{zheng2019self}, a solver referred to as \textit{self-regulated evolutionary multitasking optimization} is presented. This method introduces a \textit{self-regulated knowledge transfer scheme} that establishes several innovative concepts such as \textit{ability vector} or \textit{task-groups}. These concepts permit the amount of material shared among the population elements to be controlled.

\subsection{Cellular Genetic Algorithm} \label{sec:back_CGA}

\begin{figure}[h!]
	\centering
	\includegraphics[width=0.7\hsize]{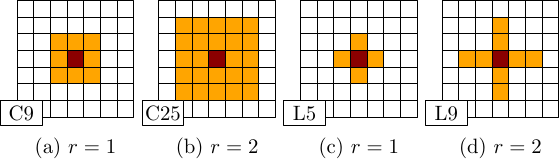}
	\caption{Examples of NEWS (a and b) and C9 (c and d) neighborhood structures, in both their canonical ($r=1$) and extended variants ($r = 2$).}
	\label{fig:structes}
\end{figure}

The herein proposed AT-MFCGA embraces cGAs at their core, which are a specific kind of Genetic Algorithm characterized by a population structured in small-sized neighborhoods \cite{manderick1989fine}. This way, each individual can only interact with solutions belonging to its neighborhood. This feature enhances the exploration of the search space through the induced slow diffusion of solutions across the population. Furthermore, exploitation is conducted within each neighborhood \cite{alba2004solving}. Hence, while classical GAs are organized in a unique panmictic population, in cGAs the population is arranged on a grid (usually two-dimensional), on which neighborhood relationships are established. Specifically, these two are the most frequently utilized neighborhood structures in cellular algorithms: i) NEWS (also referred to as linear5 or Von Neumann), in which the neighborhood of a population member is composed of its North (N), East (E), West (W), and South (S) counterparts; and ii) C9 (also known as Moore), in which the neighborhoods are composed of NW, N, NE, W, E, SW, S and SE individuals. These two structures can also be extended as shown in Figure \ref{fig:structes}, in which both canonical and extended versions of the NEWS and C9 neighborhood structures are depicted. We recommend \cite{alba2009cellular} for further information about possible cellular grid structures.

Thus, genetic operators in a cGA operate inside the neighborhood, mating each element with one of its neighbors. Additionally, newly created solutions are not merged in the whole population. Instead, they replace their previous individual upon the fulfillment of a certain criterion (usually an improvement in the fitness function). Indeed, the parallel and locally controlled interactions enabled by cellular neighborhoods have encouraged us to adopt the structure of cGAs for the algorithmic design of our proposed AT-MFCGA.

Two types of cGAs can be found in the literature depending on the policy adopted to update individuals in the grid. On the one hand, \textit{synchronous} cGAs carry out all the replacements at the same time. On the other hand, in \textit{asynchronous} cGAs individuals are sequentially updated. This approach overrides the need for any auxiliary population, so that the search can adapt faster to the newly generated genetic material. Furthermore, it naturally suits asynchronous distributed computing environments better.

\section{Proposed Adaptive Transfer-guided Multifactorial Cellular Genetic Algorithm}\label{sec:MFCGA}

Based on the background given in the previous section, we will now describe  the design of the AT-MFCGA approach proposed in this paper in depth. First, Section \ref{sec:MFCGAprel} details the non-adaptive version (MFCGA), which serves as the baseline for the design of AT-MFCGA. Next, Section \ref{sec:AT-MFCGA} elaborates on the key aspects of AT-MFCGA, focusing on its novelty and the main characteristics that make it a promising alternative when solving EM scenarios (Section \ref{sec:novelties}).

\subsection{Preliminary Steps: Multifactorial Cellular Genetic Algorithm (MFCGA)} \label{sec:MFCGAprel}

As outlined in Sections \ref{sec:MFEA} and \ref{sec:MFEA-II}, both MFEA and MFEA-II build upon four basic design principles: unified representation, assortative mating, selective evaluation, and scalar fitness based selection. These four principles are also considered when designing the MFCGA approach. This way, MFCGA must be regarded as a static MFEA counterpart that is, not sensitive to the search performance nor resilient to negative information transfer. Algorithm \ref{alg:MFCGA} shows the pseudocode of MFCGA, which is inspired by cGAs and MFEA.
\begin{algorithm}[htb]
	 \SetAlgoLined
	 \DontPrintSemicolon
		Randomly generate a population of $P$ individuals\;
		Assess each individual for all the $K$ tasks\;
		Assign the skill factor ($\tau^i$) to each member $\mathbf{x}_i$ of the population\;
		Let $\mathbf{X}_i^{\circledast}$ represent the set of neighbors of $\mathbf{x}_i$\;
		\While{termination criterion not reached}{
		    \For{$i=1,\ldots,P$}{
        		Select at random a neighbor $\mathbf{x}_j$ of $\mathbf{X}_i^{\circledast}$\; 
        		$\mathbf{x}_i^{crossover}\gets\texttt{crossover}(\mathbf{x}_i,\mathbf{x}_j)$\;
        		$\mathbf{x}_i^{mutation}\gets\texttt{mutation}(\mathbf{x}_i)$\;
    		    Evaluate $\mathbf{x}_i^{crossover}$ and $\mathbf{x}_i^{mutation}$ for $\tau^i$\;
    		    $\mathbf{x}_i \gets \texttt{best}(\mathbf{x}_i,\mathbf{x}_i^{crossover},\mathbf{x}_i^{mutation})$\;
    		    Update $\tau^i$ and $\varphi^i$ of the evolved $\mathbf{x}_i$\;
    		}
		}
		Return as solution the best individuals of each task $T_k$\;
   \caption{Pseudocode of MFCGA}
	 \label{alg:MFCGA}
\end{algorithm}

To begin with, MFCGA adopts the unified representation principle of MFEA. The genetic operators are based on the classical evolutionary crossover and mutation mechanisms. This way, at each generation these two operators are applied to each member $\mathbf{x}_i$ of the population, without any mutation or crossover probabilities. Thus, two new individuals $\mathbf{x}_i^{crossover}$ and $\mathbf{x}_i^{mutation}$ are generated. On the one hand, $\mathbf{x}_i^{crossover}$ is created as a result of mating $\mathbf{x}_i$ with a randomly selected neighbor $\mathbf{x}_j$ from the cellular neighborhood $\mathbf{X}_i^{\circledast}$ of $\mathbf{x}_i$. On the other hand, $\mathbf{x}_i^{mutation}$ is produced as a result of the application of the mutation operator to $\mathbf{x}_i$.

Once $\mathbf{x}_i^{crossover}$ and $\mathbf{x}_i^{mutation}$ have been created, they are evaluated by following the same \textit{selective evaluation} criteria described in Section \ref{sec:MFEA}. This crucial aspect ensures the computational efficiency and scalability of MFCGA. At this point we underscore that both $\mathbf{x}_i^{crossover}$ and $\mathbf{x}_i^{mutation}$ are evaluated in task $T_{\tau^i}$, where $\tau^i$ is the skill factor of $x_i$. This specific detail implies a substantial difference regarding MFEA and MFEA-II. This is due to the fact that in MFCGA one individual is devoted to optimizing the same single task throughout the entire search process, and does not change from one task to another under any circumstance. To that effect, and for the sake of the equilibrium within the population, the first evaluation and sorting is based on both factorial rank and scalar factor, thus allocating a similar number of elements to each of the considered tasks.

It is worth pausing at this point to delve into the detail of this equilibrium of the population. The reader may think that this static assignment among tasks and individuals may hinder the effective adaptability of the method throughout the search. However, AT-MFCGA overcomes this issue by introducing a dynamic rearrangement of the grid, by which individuals modify their position in the grid during the search, thereby adapting the composition of their neighborhoods and placing individuals together that are specialized in tasks with positive knowledge transfer.

Another relevant feature of the MFCGA is its \textit{local improvement selection mechanism}. Specifically, a newly generated $\mathbf{x}_i^{crossover}$ or $\mathbf{x}_i^{mutation}$ can only substitute its primal solution $\mathbf{x}_i$. Thus, the individual survivor is the best among $\mathbf{x}_i^{crossover}$, $\mathbf{x}_i^{mutation}$ and $\mathbf{x}_i$, automatically discarding the others.

\subsection{From MFCGA to AT-MFCGA: dealing with negative genetic transfer} \label{sec:AT-MFCGA}

Despite the good performance shown by MFCGA in TSP problems, it shares the same problem as the vast majority of the techniques that are designed for this very same purpose: the inability to cope efficiently with negative genetic transfers. This issue implies the execution of unprofitable crossovers that ultimately hinder convergence and penalizes the scalability of the algorithm. Therefore, we have devised new mechanisms to make MFCGA sensitive to the search process and the synergies among the tasks being solved, yielding the Adaptive Transfer-guided Multifactorial Cellular Genetic Algorithm (AT-MFCGA) that lies at the core of this research. 

The main scheme of the AT-MFCGA has been built with two main objectives in mind, which should be properly balanced:
\begin{itemize}[leftmargin=*] 
\item The first objective is the efficient adaptation of the search process to the synergies that arise from the tasks throughout the execution. This adaptation entails the addition of new additional mechanisms to MFCGA. As seen in recent studies, dealing with negative genetic transfers properly should lead to better overall results.

\item The second objective is to maintain comparable complexity levels in comparison with MFCGA and other previously published alternatives such as MFEA-II. For this reason, newly added adaptation mechanisms should be effective and computationally efficient, making the overall multitasking approach attractive and easy to implement.
\end{itemize}

After stating the two objectives above, we will now explain the two novel mechanisms: \textit{grids rebuilding} and \textit{multi-mutation} in detail. We note that these two mechanisms that embody the main proposal of this study add a small computational overhead to the original MFCGA. These two features are introduced as part of a process called \texttt{dynamicAdaptation()}, which is executed between steps 13 and 14 of Algorithm \ref{alg:MFCGA} at each $adaptiveFrequency$ generation. The pseudocode of the whole procedure is shown in Algorithm \ref{alg:dynamic}. In this pseudocode, $\texttt{isEmpty}$($\tau_{i}$) denotes a Boolean function returning a \texttt{True} value if all elements with skill factor $\tau_{i}$ have been already introduced in the new grid:
\begin{algorithm}[htb]
	\SetAlgoLined
	\DontPrintSemicolon
	\KwIn{$OldGrid$: \{$x_i,...,x_P$\}}
	\KwOut{$NewGrid$: \{$x_i,...,x_P$\}}
	\Parameter{$Rand1$: random integer $\in[1,P]$, $Rand2$: random double $\in[0,1]$, $p_{same\_task}$}
	$NewGrid$[1] = $OldGrid$[$Rand1$]\;
	Assign $\texttt{mutation}(\cdot)$ to $NewGrid$[1] through \emph{Multi-mutation} mechanism \;

	\For{$p=2,\ldots,P$}{
		
		\uIf{$Rand2<p_{same\_task}$ and not $\texttt{isEmpty}$($\tau_{p-1}$)}{
			$NewGrid$[p] = random individual from $OldGrid$ with skill factor $=\tau_{p-1}$ \;
		}\uElse{
			Assign to $\tau_k$ a value following the Roulette Wheel Selection criterion\;
			$NewGrid$[p] = random individual from $OldGrid$ with skill factor $=\tau_{k}$
		}
	
		\For{$k'=1,\ldots,K$}{
			\uIf{$\texttt{isEmpty}$($\tau_{k'}$)}{
				Recompute Roulette Wheel Selection probabilities\;
			}
		}
	
		Assign $\texttt{mutation}(\cdot)$ to $NewGrid$[p] through \emph{Multi-mutation} mechanism \;
	}
	
	\caption{\texttt{dynamicAdaptation()} mechanism of AT-MFCGA}
	\label{alg:dynamic}
\end{algorithm}

\begin{itemize}[leftmargin=*]
	
	\item \textit{Grid Rebuilding}: in most cGAs developed by the research community to date, the grid built when initializing the population remain stable throughout the algorithm execution. Although in other applications the reconstruction of this grid could seem counterproductive for the search, EM provides the perfect ground for this novel feature. Specifically, the adaptation of the grid aims to place individuals together that belong to synergistic and complementary tasks, inherently minimizing the incidence of negative genetic transfer in the search. Thus, the \textit{Grid Rebuilding} mechanism is conceived as a procedure to dynamically adapt the neighborhoods of each individual in the population.
	
	Delving now into the procedure used to rebuild the grid, AT-MFCGA records at each generation the amount of positive genetic transfers produced along the search in the form of a $K\times K$ matrix $\mathbf{G}$ of integers, where $g_{\tau_j,\tau_i}$ represents the number of times an individual $\mathbf{x}_i^{crossover}$ resulting from mating $\mathbf{x}_i$ (with skill factor $\tau_i$) and $\mathbf{x}_j$ (with skill factor $\tau_j$) outperforms $\mathbf{x}_i^{mutation}$ and $\mathbf{x}_i$. In these cases, a positive transference of genetic material has been produced, so the matrix input corresponding to the skill factors of the breeding individuals is incremented. This way, the method has an updated track of the performance of the genetic material exchange among the tasks.
	
	Every time \texttt{dynamicAdaptation()} is run, the grid is reconstructed based on the $K\times K$ matrix. This procedure begins by drawing an individual $x_i \in P$ uniformly at random, which is placed in the first position of the new grid. After this first placement, the remaining members of the population are sequentially inserted into the position adjacent to the last introduced element. For new placements, a random individual with the same skill factor $\tau_{i}$ is chosen under a probability equal to $p_{same\_task}$, which remains the same throughout the search. If learned during the search in the same fashion as the rest of the $K\times K$ matrix, the high number of mutations make $p_{same\_task}$ eventually dominate numerically over the probability of locating different tasks together. This eventually hinders the convergence of AT-MFCGA as per its cellular neighborhood structure. With probability $1-p_{same\_task}$, $\mathbf{x}_j$ is chosen by following a \textit{roulette wheel selection} procedure using the $K\times K$ matrix described above. Specifically, $\mathbf{x}_j$ is drawn at random from the individuals in the remaining population that feature skill factor $\tau_j$, where:
	\begin{equation}
	Pr(\tau_j = k)=\frac{g_{\tau_i,k}}{\sum_{k'=1}^K g_{\tau_i,k'}}.
	\end{equation}
	
	It should be noted that additional generic procedures can be found in the literature for the same purpose: to sample an individual from a ranked population as per its relative fitness value. We have used this procedure due to the fact that it is conceptually simple, yet leads to a grid rebuilding mechanism that allocates together synergistically related individuals over the cellular automata.
	
	Two crucial aspects should be considered in order to fully understanding this procedure. On the one hand, once a new row is started, the last element of the previous row is used as reference for the new insertion, so that border effects are minimized (due to e.g. rectangular cellular grids a Moore neighborhoods). On the other hand, once all the individuals with the same skill tasks are already inserted, the probabilities of the \textit{roulette wheel selection} are updated considering the tasks of individuals that are still to be deployed on the grid.
	
	\item \textit{Multi-mutation}: one of the main features of our approach is that the search process is not solely based on direct interactions among individuals. Thus, both mutation and crossover operations are granted the same importance in the search. This is the main reason for adding an adaptation mechanism related not only to the way in which individuals interact with each other, but also in the way in which solutions individually explore their nearest regions of the search space. The main contribution of this mechanism to the whole AT-MFCGA algorithm is the enhancement of the local exploration capacity of the individuals within the population. This has been carried out thanks to the dynamic variation of the neighborhood. More specifically, we modify the mutation operator of the individuals in the cellular structure so that they can explore their neighborhood using different patterns, potentially discovering better fitted solutions.
	
	To this end, AT-MFCGA is endowed with a set of mutation operators. One function is first selected from all the available functions and automatically assigned to each individual once it is created in the initialization process (step 1 of Algorithm \ref{alg:MFCGA}). After that, and as part of the newly introduced \texttt{dynamicAdaptation()} procedure, the mutation function of each individual is randomly reassigned to all the operators available, using the same probability for each function. Furthermore, within the \texttt{dynamicAdaptation()} operation, the multi-mutation mechanism is triggered after the grid rebuilding process (step 13 of Algorithm \ref{alg:dynamic}). For a better understanding of this mechanism, let us assume a possible individual $\mathbf{x}_i$, which is part of a population $P$ trying to solve a given EM environment. In doing so, AT-MFCGA contains a pool of three mutation operators $M = \{m_1,m_2, m_3\}$. At the beginning of the execution of the algorithm, an operator is selected uniformly at random from $M$ and assigned to $\mathbf{x}_i$, (e.g. $m_1$), which is used for mutation purposes (step 9 of Algorithm 3). After that, at the next time \texttt{dynamicAdaptation()} is triggered, a new mutation function is assigned to $\mathbf{x}_i$, where in our example, $m_2$ and $m_3$ are considered eligible. In other words, the mutation operator in use is forced to vary among consecutive executions of the \texttt{dynamicAdaptation()} strategy.
\end{itemize}

\begin{figure}[h!]
	\centering
	\includegraphics[width=0.9\hsize]{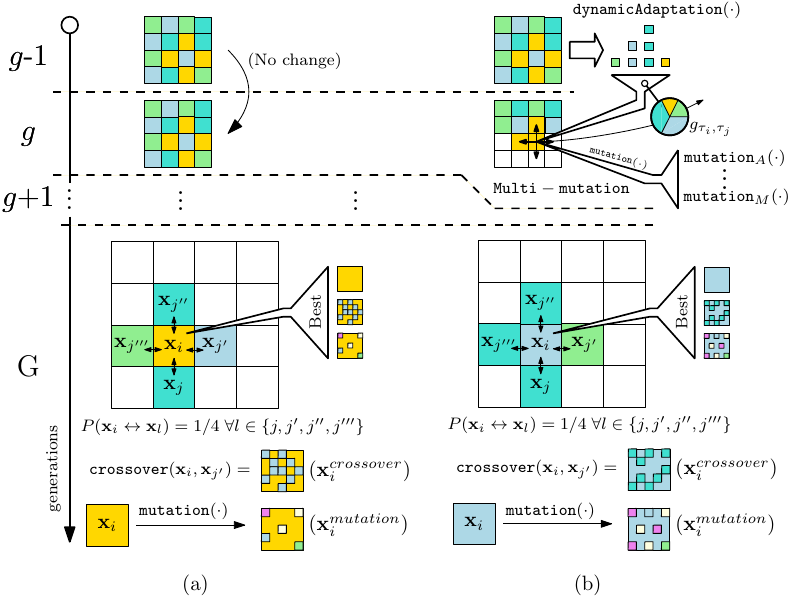}
	\caption{Graphical representation of the main differences among MFCGA (a) and AT-MFCGA (b), and how they affect the structure and behavior of the population of individuals.}
	\label{fig:differences}
\end{figure}

We finish this section by highlighting that these two simple mechanisms barely increase the computational complexity of the method, while the benefits provided to the algorithm are remarkable. Furthermore, in order to enhance the understandability of both methods, in Figure \ref{fig:differences} we have graphically shown the main differences among MFCGA and AT-MFCGA. As will be empirically shown in Section \ref{sec:exp_rebuilding}, the adaptability of AT-MFCGA to the synergies between tasks encountered throughout the search is significant and easily provable. This feature helps the solver attain better results and distribute the acquired knowledge to the population of individuals in a more efficient way.

\subsection{What makes AT-MFCGA a promising EM method?} \label{sec:novelties}

In this subsection we summarize the 4 key aspects that makes AT-MFCGA a competitive method when solving EM scenarios. It should be considered that the fulfillment of these points have acted as a guide for the design and development of this method:
\begin{itemize}[leftmargin=*]
	
	
	\item \textit{The cellular structure guarantees the scalability of the method}: the first interesting feature that characterizes AT-MFCGA is its grid-based topology used to organize the populations of individuals. The inherently parallel structure of cGAs creates an appropriate scenario for the efficient solving of EM environments, which are known to require distributed mechanisms to accommodate the huge computation effort when dealing with many optimization tasks. Moreover, the adaptation of cGA to EM opens the door to further research opportunities, by incorporating future advances in cellular algorithms into this field. These opportunities could include, for example, new distributed computing procedures for highly-dimensional problems.
	
	\item \textit{AT-MFCGA deals efficiently with negative transfers}: the most important feature of AT-MFCGA is arguably its ability to handle negative transfers between unrelated tasks. The \textit{local improvement selection} described in Section \ref{sec:MFEA} inherently minimizes the impact of negative transfers on the convergence of the solver. This is due to the fact that since our proposal generates two different individuals for each member of the population through the application of classical genetic operators. For this reason, the same importance is given to the recombination of the original individual with a neighbor, and to the local movement made through the mutation operator. In this way, for these cases in which the transfer of knowledge is not profitable, the solver inherently sets aside the individual generated through the application of the crossover function, intensifying the search on local variations of the primal solution. Furthermore, the mechanism for seeking good local solutions is improved in AT-MFCGA with the introduction of the \textit{multi-mutation} feature.
	
	\item \textit{AT-MFCGA minimizes the number of negative transfers}: AT-MFCGA implicitly downplays the importance of individuals produced as a result of the recombination process  carried out in cases in which tasks are not complementary. However, a great amount of non-profitable crossover operations could still be performed, leading to a significant waste of computational resources. These computational resources should be used to enhance the communication of synergistic tasks. As will be shown in Section \ref{sec:exp_rebuilding}, the \textit{Grid Rebuilding} mechanism further contributes to the reduction of the number of negative transfers, favoring the relationships between complementary tasks.
	
	\item \textit{Transfer-based explainability of inter-task synergies}: as pointed in the introduction, AT-MFCGA are adequate to explicitly quantify the amount of synergistic communications among the different problems being faced. This characteristic provides the researcher with an explainability interface, not only to be able to visually examine and understand the interactions between tasks, but also to help enhance the overall performance of the method with the construction of synergistic EM scenarios. This interesting feature is taken a step further by AT-MFCGA with the inclusion of the \textit{Grid Rebuilding} mechanism, whose output permits to visually inspect how individuals are reorganized within the grid according to the level of complementarity of their specialized tasks. A further discussion on this feature of AT-MFCGA is made in Section \ref{sec:exp_rebuilding}.
	
\end{itemize}

\section{Experimental Setup}\label{sec:exp}

An extensive experimental setup has been designed to assess the performance of AT-MFCGA when dealing with EM scenarios comprising different optimization tasks. Furthermore, experiments are also devised to compare the performance of AT-MFCGA to that of MFEA, MFEA-II and the non-adaptive version of our proposal (MFCGA). We also examine the genetic transfer resulting from the execution of AT-MFCGA in the considered tasks, and the influence of the synergies that appeared between these tasks on the adaptive organization of the cellular grid. 

As pointed in the introduction of this paper, the experimentation has been conducted over four different well-known combinatorial optimization problems:
\begin{itemize}[leftmargin=*]
	\item TSP  \cite{lawler1985traveling}, which is arguably one of the most recurrent problems used for benchmarking purposes and for modeling real-world logistic and transportation problems. A myriad of methods have been applied to this problem, ranging from classical solvers to more recently proposed nature-inspired techniques.
	
	\item CVRP \cite{toth2002vehicle}, which is closely related to the previous TSP, has also played a paramount role in problems arising from the transportation and mobility realms. For readers interested in the VRP problem, we recommend reference material such as \cite{osaba2020vehicle}.
	
	\item QAP \cite{lawler1963quadratic}, whose main goal is to assign a group of facilities to a set of locations, thus minimizing the total related cost. This problem has been studied in the last few decades in very different areas such as data allocation or scheduling.
	
	\item LOP \cite{bertsimas1997introduction} is the last problem considered in this experimentation. This classical combinatorial optimization problem can be briefly formulated as the search for the optimal permutation of the rows and columns of a $W\times W$ matrix $\mathbf{W}$ comprised by nonnegative values, that maximizes the accumulated value of its upper-diagonal values. This formulation allows for the modeling of a diversity of problems that arise in assorted areas such as archeology, economics or sports.
\end{itemize}

Concretely, the performance of AT-MFCGA, MFEA, MFEA-II and MFCGA has been assessed over 11 different \emph{test cases}, constructed by the combination of 20 public test instances of the above problems. To begin with, 5 TSP instances of the Krolak/Felts/Nelson benchmark comprising 100 to 200 nodes have been used, which are contained in the TSPLIB repository. In the case of the CVRP, 5 instances with 50 to 60 nodes have been considered, all of them part of the Augerat benchmark. Regarding QAP, another 5 instances with sizes 25 to 32 have been collected from the QAPLib repository. Finally, the LOP instances considered in our benchmark have been retrieved from the LOPLib library, all with a size value equal to $W=44$. 

We remark that all these 20 instances have been chosen not only due to their acceptance by the related community, but also because of the different degrees of genetic complementarities in their structure. In Table \ref{tab:similarity} we depict the genetic overlaps for the 20 instances, split by problem. These overlaps have been calculated as per the percentage of nodes (in the case of the TSP and CVRP) and weights (in the case of QAP and LOP) that the instances share. This percentage can be considered to be an early indicator of the influence of these complementarities in the positive and negative genetic exchanges of the implemented EM solvers. Two clarifications should be made at this point. First, since instances belong to different problems of different nature, we expect that negative transfers will occur in certain inter-problem interactions. Secondly, the main goal of EM algorithms is to solve different multitasking scenarios without assuming any background knowledge about the problems/tasks to be solved. This is why high-quality EM solvers aim to automatically infer positive and negative transfers by dynamically exploiting complementarities among tasks, and to attain competitive results even if such complementarities are scarce (with better outcomes in synergistic tasks).

In each of the generated test cases, the implemented solvers should tackle the tasks belonging to the test case simultaneously. Table \ref{tab:testCases} summarizes the composition of each of these test cases. As can be seen, we have first constructed a test case for each considered problem (TSP, CVRP, QAP and LOP), comprising the 5 instances of each case. Then we have constructed 6 medium-sized test cases, each comprising 10 different instances of two problems (5 instances per problem). Finally, we have arranged a large test case considering all the 20 problem instances under consideration. The reasons for performing experiments with these 11 test cases is twofold: i) to increase the heterogeneity and variety of configurations and problems being solved; and ii) to assess whether the complementarities represented in Table \ref{tab:similarity} can be exploited even if instances belong to different optimization problems. To the best of our knowledge, this is the largest discrete multitasking environment solved by Transfer Optimization. 
\begin{table}[ht!]
	\centering
	\caption{Summary of genetic complementarities for all the instances employed in the experimentation, computed as the percentage of overlap between the nodes/weights of every pair of instances in comparison}	
	\resizebox{0.8\columnwidth}{!}{
		\begin{tabular}{cccccc}
			\toprule
			Instance & kroA100 & kroA150 & kroA200 &  kroB150 & kroC100\\
			\midrule
			kroA100 & \cellcolor{gray!25} & \textbf{80\%} &  \textbf{66\%} & 1\% & 2\%\\
			kroA150 & \cellcolor{gray!25} & \cellcolor{gray!25} & \textbf{57\%} & 1\% & 1\%\\
			kroA200 & \cellcolor{gray!25} & \cellcolor{gray!25} & \cellcolor{gray!25} & \textbf{66\%} & \textbf{57\%}\\ 
			kroB150 & \cellcolor{gray!25} & \cellcolor{gray!25} & \cellcolor{gray!25} & \cellcolor{gray!25} & \textbf{80\%}\\
			\toprule
			Instance & P-n50-k7 & P-n50-k8 & P-n55-k7, & P-n55-k15 & P-n60-k10\\
			\midrule
			P-n50-k7 & \cellcolor{gray!25} & \textbf{100\%} &  \textbf{95\%} & \textbf{95\%} & \textbf{90\%}\\
			P-n50-k8 & \cellcolor{gray!25} & \cellcolor{gray!25} & \textbf{95\%} & \textbf{95\%} & \textbf{90\%}\\
			P-n55-k7 & \cellcolor{gray!25} & \cellcolor{gray!25} & \cellcolor{gray!25} & \textbf{100\%} & \textbf{95\%}\\ 
			P-n55-k15 & \cellcolor{gray!25} & \cellcolor{gray!25} & \cellcolor{gray!25} & \cellcolor{gray!25} & \textbf{100\%}\\
			\toprule
			Instance & Nug25 & Nug30 & Kra30a & Kro30b & Kra32\\
			\midrule
			Nug25 & \cellcolor{gray!25} & \textbf{35\%} &  0\% & 0\% & 0\%\\
			Nug30 & \cellcolor{gray!25} & \cellcolor{gray!25} & 0\% & 0\% & 0\% \\
			Kro30a & \cellcolor{gray!25} & \cellcolor{gray!25} & \cellcolor{gray!25} & \textbf{50\%} & \textbf{25\%}\\ 
			Kro30b & \cellcolor{gray!25} & \cellcolor{gray!25} & \cellcolor{gray!25} & \cellcolor{gray!25} & \textbf{37\%}\\
			\toprule
			Instance & N-t59d11xx & N-t59f11xx & N-t59i11xx & N-t65f11xx & N-t70f11xx\\
			\midrule
			N-t59d11xx & \cellcolor{gray!25} & 2\% &  \textbf{26\%} & \textbf{16\%} & \textbf{17\%}\\
			N-t59f11xx & \cellcolor{gray!25} & \cellcolor{gray!25} & \textbf{16\%} & \textbf{10\%} & 8\% \\
			N-t59i11xx & \cellcolor{gray!25} & \cellcolor{gray!25} & \cellcolor{gray!25} & 10\% & \textbf{16\%}\\ 
			N-t65f11xx & \cellcolor{gray!25} & \cellcolor{gray!25} & \cellcolor{gray!25} & \cellcolor{gray!25} & \textbf{10\%}\\
			\bottomrule
		\end{tabular}
	}
	\label{tab:similarity}
\end{table}

Regarding the parameters used for each solver, and for the sake of a fair comparison we have chosen similar values and operators for all the techniques employed. Furthermore, we ensure the reproducibility of the presented results by showing the configuration of all the considered methods in Table \ref{tab:Parametrization1}. Regarding AT-MFCGA, the parameters listed in Table \ref{tab:Parametrization1} are used as the configuration of the algorithm. Additionally, in order to define the values of these and other parameters in the experimentation, we have focused on the guidelines and configurations previously reported in studies dealing with cGAs, MFEA and MFEA-II for discrete optimization problems \cite{alba2004solving,yuan2016evolutionary,osaba2020dmfea}. Further verification tests have been conducted to ensure that all these parameters gave rise to good performance in all the algorithms. This trend has also been followed for the methods used in the comparison: MFEA, MFEA-II and MFCGA. Additionally, good methodological practices for bio-inspired optimization research have also been embraced \cite{latorre2020fairness}: each test case has been run $20$ times, and hypothesis tests have been applied to obtain informed insight on the statistical significance of the reported performance gaps. Furthermore, as termination criteria, all EM methods used in this experimentation end their execution after $500\cdot 10^3$ objective function evaluations.
	
Finally, in order to take a step beyond the replicability of this experimentation, a public JAVA implementation of AT-MFCGA has been made publicly available at the following repository: \url{https://git.code.tecnalia.com/aritz.martinez/at-mfcga}. This repository also contains the source code that produces the results presented and discussed in this paper. Furthermore, interested readers can also find the source code of the basic version of the method, MFCGA, in: \url{https://git.code.tecnalia.com/aritz.martinez/mfcga}.

\begin{table}[ht!]
	\centering
	\caption{Summary of the 11 test cases built for the experimentation.}
	\resizebox{0.8\columnwidth}{!}{
		\begin{tabular}{cL{12cm}}
			\toprule
			Test Case & Instances involved\\ \midrule
			\texttt{TC\_TSP} & \texttt{kroA100}, \texttt{kroA150}, \texttt{kroA200}, \texttt{kroB150}, \texttt{kroC100}\\ 
			\texttt{TC\_VRP} & \texttt{P-n50-k7}, \texttt{P-n50-k8}, \texttt{P-n55-k7}, \texttt{P-n55-k15}, \texttt{P-n60-k10}\\ 
			\texttt{TC\_QAP} & \texttt{Nug25}, \texttt{Nug30}, \texttt{Kra30a}, \texttt{Kro30b}, \texttt{Kra32}\\ 
			\texttt{TC\_LOP} & \texttt{N-t59d11xx}, \texttt{N-t59f11xx}, \texttt{N-t59i11xx}, \texttt{N-65f11xx}, \texttt{N-70f11xx}\\ 
			\texttt{TC\_TSP\_VRP} & \texttt{TC\_TSP} $\cup$  \texttt{TC\_VRP}\\ 
			\texttt{TC\_TSP\_QAP} & \texttt{TC\_TSP} $\cup$  \texttt{TC\_QAP}\\ 
			\texttt{TC\_TSP\_LOP} & \texttt{TC\_TSP} $\cup$  \texttt{TC\_LOP}\\ 
			\texttt{TC\_VRP\_QAP} & \texttt{TC\_VRP} $\cup$  \texttt{TC\_QAP}\\ 
			\texttt{TC\_VRP\_LOP} & \texttt{TC\_VRP} $\cup$  \texttt{TC\_LOP}\\ 
			\texttt{TC\_QAP\_LOP} & \texttt{TC\_QAP} $\cup$  \texttt{TC\_LOP}\\ 
			\texttt{TC\_ALL} &  \texttt{TC\_TSP} $\cup$  \texttt{TC\_VRP} $\cup$  \texttt{TC\_QAP} $\cup$  \texttt{TC\_LOP}\\ \bottomrule
		\end{tabular}
	}
	\label{tab:testCases}
\end{table}

\begin{table*}[h!]
	\centering
	\caption{Parameter values for MFEA, MFEA-II, MFCGA and AT-MFCGA.}
	\label{tab:Parametrization1}
	\resizebox{\columnwidth}{!}{
		\begin{tabular}{lcclcclcclcc}
			\toprule 
			\multicolumn{2}{c}{MFEA} & & \multicolumn{2}{c}{MFEA-II} & & \multicolumn{2}{c}{MFCGA} & & \multicolumn{2}{c}{AT-MFCGA} \\
			\cmidrule{1-2} \cmidrule{4-5} \cmidrule{7-8} \cmidrule{10-11}
			Parameter & Value & & Parameter & Value & & Parameter & Value & & Parameter & Value\\
			\cmidrule{1-2} \cmidrule{4-5}  \cmidrule{7-8} \cmidrule{10-11}
			$P$ size (small TCs) & 200 & & $P$ size (small TCs) & 200 & & Grid size (small TCs) & $10\times 20$ & & Grid size (small TCs) & $10\times 20$ \\ 
			$P$ size (large TCs) & 300 & & $P$ size (large TCs) & 300 & & Grid size (large TCs) & $10\times 30$ & & Grid size (large TCs) & $10\times 30$ \\
			$\texttt{CX}$ & OX \cite{davis1985job} & & $\texttt{Intra-task CX}(\cdot)$ & OX & & $\texttt{CX}(\cdot)$ & OX & & $\texttt{CX}(\cdot)$ & OX \\ 
			$\texttt{mutation}(\cdot)$ & 2-opt \cite{lin1965computer} & & $\texttt{mutation}(\cdot)$ & 2-opt & & $\texttt{mutation}(\cdot)$ & 2-opt & & $\texttt{mutation}(\cdot)$ & \{2-opt, Insertion\} \\
			RMP & 0.9 & & Initial values of $RMP_{k,k'}$ & 0.95 & & Neighborhood & Moore & & Neighborhood & Moore \\
			& & & $\texttt{Parent Centric CX}(\cdot)$ & Dynamic OX & & & & & $adaptiveFrequency$ & 100 generations\\
			& & & $P_m$ & 0.2 & & & & & $p_{same\_task}$ & 0.5\\
			& & & $\Delta_{inc}$ / $\Delta_{dec}$ & 0.99 / 0.99 & &  & & \\\bottomrule
		\end{tabular}
	}
\end{table*}

\section{Results and Discussion}\label{sec:exp_res}

Table \ref{tab:results} summarizes the results obtained by MFEA, MFEA-II, MFCGA and AT-MFCGA for all the designed test cases. Specifically, each entry of this table indicates the average and standard deviation of the fitness attained for each instance and test cases, computed over the $20$ independent runs performed for every test case. Best average results have been highlighted in bold to ease their visual inspection. We also provide the optima for each instance that has been given by the repositories from where the instances have been retrieved. Nevertheless, it is important to point out that the objective of this experimentation is not to reach the optimal solution of every instance, but rather to use these solutions as references of the performance of the compared multitasking methods.
\begin{table*}[h!]
	\centering
	\caption{Results obtained by MFEA, MFEA-II, MFCGA and AT-MFCGA for all the test environments. Best average results have been highlighted in bold. Each (algorithm,instance) entry indicates the mean (top) and standard deviation (bottom) of the fitness over 20 runs. Fitness values of LOP instances are negative to invert the direction of the search (maximize $\mapsto$ minimize).}
	\renewcommand{\arraystretch}{0.92}
	\resizebox{0.92\columnwidth}{!}{
		\begin{tabular}{cccccccccccccccccccccc}
			\toprule
			& \multirow{2}{*}{Method} & \multicolumn{5}{c}{TSP instances} & \multicolumn{5}{c}{CVRP instances} & \multicolumn{5}{c}{QAP instances} & \multicolumn{5}{c}{LOP instances}\\
			\cmidrule(lr){3-7} \cmidrule(lr){8-12} \cmidrule(lr){13-17} \cmidrule(lr){18-22}
			& & \rotatebox{90}{\texttt{kroA100}} & \rotatebox{90}{\texttt{kroA150}} & \rotatebox{90}{\texttt{kroA200}} & \rotatebox{90}{\texttt{kroB150}} & \rotatebox{90}{\texttt{kroC100}} & \rotatebox{90}{\texttt{P-n50-k7}} & \rotatebox{90}{\texttt{P-n50-k8}} & \rotatebox{90}{\texttt{P-n55-k7}} & \rotatebox{90}{\texttt{P-n55-k15}} & \rotatebox{90}{\texttt{P-n60-k10}} & \rotatebox{90}{\texttt{Nug25}} & \rotatebox{90}{\texttt{Nug30}} & \rotatebox{90}{\texttt{Kra30a}} & \rotatebox{90}{\texttt{Kra30b}} & \rotatebox{90}{\texttt{Kra32}} & \rotatebox{90}{\texttt{N-t59d11xx}} & \rotatebox{90}{\texttt{N-t59f11xx}} & \rotatebox{90}{\texttt{N-t59i11xx}} & \rotatebox{90}{\texttt{N-65f11xx}} & \rotatebox{90}{\texttt{N-70f11xx}}\\ \midrule
			\multirow{9}{*}{\rotatebox{90}{\emph{TC\_TSP}}} & 
			\multirow{2}{*}{MFEA} 
			& 22925.0 & 31127.1 & 33694.5 & 31601.3 & 23199.2 & -- & -- & -- & -- & -- & -- & -- & -- & -- & -- & -- & -- & -- & -- & --\\ 
			& & 845.55 & 554.62 & 737.29 & 996.60 & 619.85 & -- & -- & -- & -- & -- & -- & -- & -- & -- & -- & -- & -- & -- & -- & --\\ 
			\cmidrule{2-22}
			& \multirow{2}{*}{MFEA-II} 
			& 22419.9 & 29432.8 & 32097.8 & 28601.5 & 22604.4 & -- & -- & -- & -- & --  & -- & -- & -- & -- & -- & -- & -- & -- & -- & --\\ 
			& & 564.91 & 234.15 & 589.14 & 622.52 & 250.29 & -- & -- & -- & -- & --  & -- & -- & -- & -- & -- & -- & -- & -- & -- & --\\  
			\cmidrule{2-22}
			& \multirow{2}{*}{MFCGA} 
			& 21950.6 & 28383.4 & 31710.5 & 27717.5 & 21506.1 & -- & -- & -- & -- & -- & -- & -- & -- & -- & -- & -- & -- & -- & -- & --\\ 
			& & 226.34 & 372.35 & 426.96 & 499.32 & 283.46 & -- & -- & -- & -- & -- & -- & -- & -- & -- & -- & -- & -- & -- & -- & --\\ 
			\cmidrule{2-22}
			& \multirow{2}{*}{AT-MFCGA} 
			& \textbf{21883.8} & \textbf{28057.9} & \textbf{31196.9} & \textbf{27430.4} & \textbf{21411.5} & -- & -- & -- & -- & -- & -- & -- & -- & -- & -- & -- & -- & -- & -- & --\\ 
			& & 417.13 & 449.85 & 543.64 & 365.42 & 255.33 & -- & -- & -- & -- & -- & -- & -- & -- & -- & -- & -- & -- & -- & -- & --\\  
			\midrule
			\multirow{9}{*}{\rotatebox{90}{\emph{TC\_VRP}}} & 
			\multirow{2}{*}{MFEA} 
			& -- & -- & -- & -- & -- & 623.8 & 701.1 & 665.3 & 1011.9 & 840.4 & -- & -- & -- & -- & -- & -- & -- & -- & -- & --\\ 
			& & -- & -- & -- & -- & -- & 17.35 & 22.20 & 21.01 & 20.16 & 40.65 & -- & -- & -- & -- & -- & -- & -- & -- & -- & --\\ 
			\cmidrule{2-22}
			& \multirow{2}{*}{MFEA-II} 
			& -- & -- & -- & -- & -- & 602.5 & 680.2 & 673.0 & 989.3 & 819.9 & -- & -- & -- & -- & -- & -- & -- & -- & -- & --\\ 
			& & -- & -- & -- & -- & -- & 25.76 & 20.72 & 29.74 & 24.62 & 28.12 & -- & -- & -- & -- & -- & -- & -- & -- & -- & --\\  
			\cmidrule{2-22}
			& \multirow{2}{*}{MFCGA} 
			& -- & -- & -- & -- & -- & 586.9 & 669.5 & 610.0 & 995.7 & 805.5 & -- & -- & -- & -- & -- & -- & -- & -- & -- & --\\ 
			& & -- & -- & -- & -- & -- & 13.50 & 8.50 & 7.15 & 11.71 & 11.72 & -- & -- & -- & -- & -- & -- & -- & -- & -- & --\\ 
			\cmidrule{2-22}
			& \multirow{2}{*}{AT-MFCGA} 
			& -- & -- & -- & -- & -- & \textbf{583.2} & \textbf{664.0} & \textbf{604.6} & \textbf{984.2} & \textbf{797.5} & -- & -- & -- & -- & -- & -- & -- & -- & -- & --\\ 
			& & -- & -- & -- & -- & -- & 11.58 & 12.91 & 13.63 & 16.79 & 12.45 & -- & -- & -- & -- & -- & -- & -- & -- & -- & --\\  
			\midrule	
			\multirow{9}{*}{\rotatebox{90}{\emph{TC\_QAP}}} & 
			\multirow{2}{*}{MFEA} 
			& -- & -- & -- & -- & -- & -- & -- & -- & -- & -- & 4068.8 & 6768.8 & 101321.0 & 101265.0 & 99416.0 & -- & -- & -- & -- & -- \\ 
			& & -- & -- & -- & -- & -- & -- & -- & -- & -- & -- & 69.83 & 120.53 & 3393.98 & 2973.0 & 2827.21 & -- & -- & -- & -- & -- \\  
			\cmidrule{2-22}
			& \multirow{2}{*}{MFEA-II} 
			& -- & -- & -- & -- & -- & -- & -- & -- & -- & -- & 4107.0 & 6646.6 & 97616.0 & 98797.0 & 96843.0 & -- & -- & -- & -- & -- \\ 
			& & -- & -- & -- & -- & -- & -- & -- & -- & -- & -- & 85.31 & 85.57 & 1720.83 & 2932.02 & 2055.47 & -- & -- & -- & -- & -- \\ 
			\cmidrule{2-22}
			& \multirow{2}{*}{MFCGA} 
			& -- & -- & -- & -- & -- & -- & -- & -- & -- & -- & 3964.9 & 6573.2 & 95721.5 & 96806.0 & 95396.5 & -- & -- & -- & -- & -- \\ 
			& & -- & -- & -- & -- & -- & -- & -- & -- & -- & -- & 48.28 & 57.16 & 1049.83 & 1062.55 & 1431.35 & -- & -- & -- & -- & -- \\  
			\cmidrule{2-22}
			& \multirow{2}{*}{AT-MFCGA} 
			& -- & -- & -- & -- & -- & -- & -- & -- & -- & -- & \textbf{3950.0} & \textbf{6564.6} & \textbf{95535.5} & \textbf{96383.0} & \textbf{95179.0} & -- & -- & -- & -- & -- \\ 
			& & -- & -- & -- & -- & -- & -- & -- & -- & -- & -- & 58.59& 43.55 & 1313.23 & 1540.23 & 1296.81 & -- & -- & -- & -- & -- \\   
			\midrule
			\multirow{9}{*}{\rotatebox{90}{\emph{TC\_LOP}}} & 
			\multirow{2}{*}{MFEA} 
			& -- & -- & -- & -- & -- & -- & -- & -- & -- & -- & -- & -- & -- & -- & -- & -141930.6 & -120577.3 & -8149786.0 & -214448.8 & -352931\\ 
			& & -- & -- & -- & -- & -- & -- & -- & -- & -- & -- & -- & -- & -- & -- & -- & 2595.52& 2477.61 & 938.15 & 66574.67 & 2206.62\\  
			\cmidrule{2-22}
			& \multirow{2}{*}{MFEA-II} 
			& -- & -- & -- & -- & -- & -- & -- & -- & -- & -- & -- & -- & -- & -- & -- & -143557.1 & -121683.4 & -8204880.6 & -215744.8 & -354715.1 \\ 
			& & -- & -- & -- & -- & -- & -- & -- & -- & -- & -- & -- & -- & -- & -- & -- & 1427.67 & 493.70 & 40069.50 & 1265.54 & 2595.52 \\
			\cmidrule{2-22}
			& \multirow{2}{*}{MFCGA} 
			& -- & -- & -- & -- & -- & -- & -- & -- & -- & -- & -- & -- & -- & -- & -- & -145677.2 & -122284.4 & -8246649.7 & -215087.3 & -356379.9\\ 
			& & -- & -- & -- & -- & -- & -- & -- & -- & -- & -- & -- & -- & -- & -- & -- & 1100.62& 227.88 & 7901.64 & 351.30 & 1164.44\\  
			\cmidrule{2-22}
			& \multirow{2}{*}{AT-MFCGA} 
			& -- & -- & -- & -- & -- & -- & -- & -- & -- & -- & -- & -- & -- & -- & -- & \textbf{-147024.4} & \textbf{-122518.8} & \textbf{-8261545} & \textbf{-216849.1} & \textbf{-359361.0}\\ 
			& & -- & -- & -- & -- & -- & -- & -- & -- & -- & -- & -- & -- & -- & -- & -- & 363.93& 366.58 & 6590.01 & 171.54 & 399.06\\   
			\midrule
			\multirow{9}{*}{\rotatebox{90}{\emph{TC\_TSP\_VRP}}} & 
			\multirow{2}{*}{MFEA} 
			& 22973.2 & 32004.1 & 33702.1 & 31402.9 & 23208.7  & 617.2 & 718.5 & 658.7 & 1004.0 & 832.0 & -- & -- & -- & -- & -- & -- & -- & -- & -- & --\\ 
			& & 903.88 & 565.82 & 750.32 & 991.95 & 600.35 & 21.53 & 25.92 & 30.05 & 25.01 & 77.48 & -- & -- & -- & -- & -- & -- & -- & -- & -- & --\\ 
			\cmidrule{2-22}
			& \multirow{2}{*}{MFEA-II} 
			& 22215.3 & 29730.7 & 32050.4 & 28700.1 & 22532.4  & 596.3 & 672.4 & 680.3 & 992.4 & 822.6 & -- & -- & -- & -- & -- & -- & -- & -- & -- & --\\ 
			& & 610.21 & 381.35 & 612.72 & 700.95 & 357.64 & 32.74 & 25.99 & 40.21 & 28.01 & 36.83 & -- & -- & -- & -- & -- & -- & -- & -- & -- & --\\   
			\cmidrule{2-22}
			& \multirow{2}{*}{MFCGA} 
			& 21902.2 & 28290.9 & 31902.9 & 27702.2 & 21603.9 & \textbf{591.7} & \textbf{669.4} & 616.5 & 996.7 & \textbf{806.1} & -- & -- & -- & -- & -- & -- & -- & -- & -- & --\\ 
			& & 230.33 & 365.79 & 462.33 & 259.94 & 269.74 & 10.00 & 8.07 & 10.47 & 15.31 & 9.55 & -- & -- & -- & -- & -- & -- & -- & -- & -- & --\\ 
			\cmidrule{2-22}
			& \multirow{2}{*}{AT-MFCGA} 
			& \textbf{21841.9} & \textbf{27864.2} & \textbf{31186.4} & \textbf{27430.4} & \textbf{21491.1} & 593.4 & 671.6 & \textbf{615.2} & \textbf{985.8} & 814.3 & -- & -- & -- & -- & -- & -- & -- & -- & -- & --\\ 
			& & 374.94 & 554.74 & 332.57 & 469.01 & 406.38 & 16.40 & 12.53 & 19.18 & 15.31 & 21.13 & -- & -- & -- & -- & -- & -- & -- & -- & -- & --\\  
			\midrule	
			\multirow{9}{*}{\rotatebox{90}{\emph{TC\_TSP\_QAP}}} & 
			\multirow{2}{*}{MFEA} 
			& 22815.5 & 30491.2 & 32749.6 & 31017.0 & 23291.5 & -- & -- & -- & -- & -- & 4170.5 & 6814.3 & 100819.4 & 102031.9 & 100800.0 & -- & -- & -- & -- & --\\ 
			& & 955.76 & 703.77 & 819.39 & 1013.95 & 690.32 & -- & -- & -- & -- & -- & 100.83 & 100.21 & 3806.42 & 3850.01 & 3603.90 & -- & -- & -- & -- & --\\ 
			\cmidrule{2-22}
			& \multirow{2}{*}{MFEA-II} 
			& 22450.1 & 29600.6 & 31892.0 & 28842.4 & 22669.1   & -- & -- & -- & -- & -- & 4032.4 & 6752.3 & 980689.2 & 98512.0 & 97076.9 & -- & -- & -- & -- & --\\ 
			& & 600.84 & 315.04 & 632.83 & 680.94 & 244.53 & -- & -- & -- & -- & -- & 78.02 & 50.01 & 3021.43 & 2599.97 & 2600.79 & -- & -- & -- & -- & --\\  
			\cmidrule{2-22}
			& \multirow{2}{*}{MFCGA} 
			& 22031.0 & 28369.4 & 31980.9 & 27747.4 & 21504.7 & -- & -- & -- & -- & -- & \textbf{3962.9} & 6581.3 & 97127.5 & 98261.5 & 97534.0 & -- & -- & -- & -- & --\\ 
			& & 238.53 & 413.06 & 494.39 & 444.52 & 328.16 & -- & -- & -- & -- & -- & 41.47 & 34.29 & 1906.44 & 1561.78 & 2229.12 & -- & -- & -- & -- & --\\ 
			\cmidrule{2-22}
			& \multirow{2}{*}{AT-MFCGA} 
			& \textbf{21911.2} & \textbf{27973.0} & \textbf{31273.7} & \textbf{27654.3} & \textbf{21460.2} & -- & -- & -- & -- & -- & 3982.0 & \textbf{6574.6} & \textbf{97067.5} & \textbf{98310.5} & \textbf{96699.5} & -- & -- & -- & -- & --\\ 
			& & 407.6 & 447.3 & 553.91 & 481.15 & 403.5  & -- & -- & -- & -- & -- & 39.74 & 63.04 & 2196.5 & 1809.3 & 2235.6 & -- & -- & -- & -- & --\\   
			\midrule
			\multirow{9}{*}{\rotatebox{90}{\emph{TC\_TSP\_LOP}}} & 
			\multirow{2}{*}{MFEA} 
			& 22867.2 & 30574.8 & 32739.0 & 31486.0 & 23280.2  & -- & -- & -- & -- & -- & -- & -- & -- & -- & -- & -139119.9 & -118590.2 & -7968085.5 & -207933.2 & -342932.2 \\ 
			& & 949.01 & 597.14 & 992.44 & 950.43 & 700.99 & -- & -- & -- & -- & -- & -- & -- & -- & -- & -- & 2313.81 & 1876.06 & 4302.32 & 3140.17 & 5540.02 \\ 
			\cmidrule{2-22}
			& \multirow{2}{*}{MFEA-II} 
			& 22152.0 & 29229.1 & 31928.3 & 28602.0 & 22792.2  & -- & -- & -- & -- & -- & -- & -- & -- & -- & -- & -143794.6 & -121998.4 & -8240106.5 & -215190.2 & -356432.8\\ 
			& & 588.64 & 270.93 & 700.79 & 682.14 & 301.02 & -- & -- & -- & -- & -- & -- & -- & -- & -- & -- & 1070.06 & 212.10 & 2878.09 & 592.33 & 3007.18\\  
			\cmidrule{2-22}
			& \multirow{2}{*}{MFCGA} 
			& 21941.1 & 28351.9 & 31966.1 & 27778.4 & 21561.1 & -- & -- & -- & -- & -- & -- & -- & -- & -- & -- & -145014.6 & -121995.2 & -8232359.4 & 216008.8 & -356234.7 \\ 
			& & 296.55 & 428.54 & 473.75 & 420.46 & 323.44 & -- & -- & -- & -- & -- & -- & -- & -- & -- & -- & 771.21 & 230.98 & 2197.02 & 289.84 & 739.35\\ 
			\cmidrule{2-22}
			& \multirow{2}{*}{AT-MFCGA} 
			& \textbf{21845.7} & \textbf{27763.1} & \textbf{31195.0} & \textbf{27464.1} & \textbf{21508.6} & -- & -- & -- & -- & -- & -- & -- & -- & -- & -- & \textbf{-147142.7} & \textbf{-122517.2} & \textbf{-8261545.0} & \textbf{-216869.7} & \textbf{-359468.2} \\ 
			& & 383.47 & 376.47 & 520.82 & 443.82 & 403.5  & -- & -- & -- & -- & -- & -- & -- & -- & -- & -- & 259.59 & 109.59 & 1100.12 & 228.54 & 473.29\\    	
			\midrule
			\multirow{9}{*}{\rotatebox{90}{\emph{TC\_VRP\_QAP}}} & 
			\multirow{2}{*}{MFEA} 
			& -- & -- & -- & -- & -- & 642.3 & 709.2 & 665.2 & 1034.5 & 870.2 & 4130.0 & 6803.4 & 100718.0 & 101928.0 & 100733.0 & -- & -- & -- & -- & --\\ 
			& & -- & -- & -- & -- & -- & 24.2 & 34.83 & 36.39 & 41.23 & 51.91 & 50.78 & 86.63 & 3798.44 & 3759.36 & 3284.81 & -- & -- & -- & -- & --\\ 
			\cmidrule{2-22}
			& \multirow{2}{*}{MFEA-II} 
			& -- & -- & -- & -- & -- & 632.9 & 703.8 & 657.5 & 1014.9 & 867.6 & 4066.6 & 6733.4 & 98268.0 & 98389.0 & 98015.0  & -- & -- & -- & -- & --\\ 
			& & -- & -- & -- & -- & -- & 19.16 & 25.56 & 30.65 & 25.62 & 46.85 & 67.74 & 39.03 & 2891.13 & 2325.8 & 2419.48 & -- & -- & -- & -- & --\\  
			\cmidrule{2-22}
			& \multirow{2}{*}{MFCGA} 
			& -- & -- & -- & -- & -- & 592.2 & 671.9 & 616.6 & 995.5 & 821.7 & 3968.3 & 6597.5 & 97793.0 & 98703.5 & 98277.0 & -- & -- & -- & -- & --\\ 
			& & -- & -- & -- & -- & -- & 8.93 & 13.60 & 11.20 & 13.61 & 14.93 & 40.64 & 32.48 & 1473.90 & 1518.72 & 1318.72 & -- & -- & -- & -- & --\\ 
			\cmidrule{2-22}
			& \multirow{2}{*}{AT-MFCGA} 
			& -- & -- & -- & -- & -- & \textbf{586.0} & \textbf{661.2} & \textbf{612.4} & \textbf{993.4} & \textbf{801.2} & \textbf{3959.9} & \textbf{6573.}0 & \textbf{96637.0} & \textbf{97716.5} &\textbf{ 96291.9} & -- & -- & -- & -- & --\\ 
			& & -- & -- & -- & -- & -- & 14.79 & 9.05 & 16.2 & 17.38 & 13.86 & 41.21 & 62.71 & 1639.58 & 1716.33 & 1770.0 & -- & -- & -- & -- & --\\   	
			\midrule
			\multirow{9}{*}{\rotatebox{90}{\emph{TC\_VRP\_LOP}}} & 
			\multirow{2}{*}{MFEA} 
			& -- & -- & -- & -- & -- & 647.2 & 710.6 & 689.2 & 1125.6 & 893.8 & -- & -- & -- & -- & -- & -140768.5 & -118738.2 & -8065891.1 & -211554.2 & -349389.1 \\ 
			& & -- & -- & -- & -- & -- & 38.23 & 28.52 & 38.80 & 29.82 & 42.91 & -- & -- & -- & -- & -- & 3144.10 & 1635.78 & 4801.40 & 1806.31 & 3419.38 \\ 
			\cmidrule{2-22}
			& \multirow{2}{*}{MFEA-II} 
			& -- & -- & -- & -- & -- & 625.9 & 704.8 & 671.6 & 1049.7 & 887.8 & -- & -- & -- & -- & --  & -143997.0 & -121710.9 & -8202345.3 & -215237.6 & -354737.7\\ 
			& & -- & -- & -- & -- & -- & 24.59 & 15.49 & 31.47 & 16.69 & 27.26 & -- & -- & -- & -- & -- & 1300.87 & 564.26 & 2748.98 & 690.58 & 1565.91\\  
			\cmidrule{2-22}
			& \multirow{2}{*}{MFCGA} 
			& -- & -- & -- & -- & -- & 590.05 & 670.5 & 615.2 & 997.1 & 804.1 & -- & -- & -- & -- & --  & -144619.0 & -122086.5 & -8237236.2 & 215899.2 & -355770.0 \\ 
			& & -- & -- & -- & -- & -- & 12.65 & 9.72 & 10.70 & 17,14 & 12.32 & -- & -- & -- & -- & -- & 846.31 & 191.16 & 2717.05 & 281.26 & 923.31\\ 
			\cmidrule{2-22}
			& \multirow{2}{*}{AT-MFCGA} 
			& -- & -- & -- & -- & -- & \textbf{583.5} & \textbf{662.3 }& \textbf{604.6} & \textbf{986.6} & \textbf{798.8} & -- & -- & -- & -- & --  & \textbf{-147258.1} & \textbf{-122519.4} & \textbf{-8261545.0} & \textbf{-216897.7} & \textbf{-359534.3} \\ 
			& & -- & -- & -- & -- & -- & 13.74 & 13.24 & 12.08 & 19.06 & 12.06 & -- & -- & -- & -- & -- & 233.77 & 236.73 & 2141.0 & 220.71 & 496.28\\     
			\midrule
			\multirow{9}{*}{\rotatebox{90}{\emph{TC\_QAP\_LOP}}} & 
			\multirow{2}{*}{MFEA} 
			& -- & -- & -- & -- & -- & -- & -- & -- & -- & -- & 4103.0 & 6796.0 & 101677.0 & 101508.0 & 100857.0 & -140768.5 & -118738.2 & -8065891.1 & -211554.2 & -349389.1 \\ 
			& & -- & -- & -- & -- & -- & -- & -- & -- & -- & -- & 68.95 & 76.05 & 2399.80 & 2031.24 & 2815.81 & 3144.10 & 1635.78 & 4801.40 & 1806.31 & 3419.38 \\  
			\cmidrule{2-22}			
			& \multirow{2}{*}{MFEA-II} 
			& -- & -- & -- & -- & -- & -- & -- & -- & -- & -- & 4040.8 & 6715.8 & 99111.0 & 99477.0 & 99032.0 & -140900.6 & -120095.6 & -8093245.7 & -213327.4 & -352395.5 \\  
			& & -- & -- & -- & -- & -- & -- & -- & -- & -- & -- & 51.90 & 74.90 & 2116.49 & 1849.21 & 2147.79 & 1367.32 & 1635.78 & 3835.75 & 1021.93 & 1542.24 \\  
			\cmidrule{2-22}
			& \multirow{2}{*}{MFCGA} 
			& -- & -- & -- & -- & -- & -- & -- & -- & -- & -- & \textbf{3976.2} & 6595.3 & 97977.5 & 98950.0 & 98414.0 & -140768.5 & -118738.2 & -8065891.1 & -211554.2 & -349389.1 \\ 
			& & -- & -- & -- & -- & -- & -- & -- & -- & -- & -- & 46.63 & 55.32 & 1210.83 & 1417.24 & 1203.70 & 3144.10 & 1635.78 & 4801.40 & 1806.31 & 3419.38 \\
			\cmidrule{2-22}
			& \multirow{2}{*}{AT-MFCGA} 
			& -- & -- & -- & -- & -- & -- & -- & -- & -- & -- & 3976.8 & \textbf{6546.9} & \textbf{96767.0} & \textbf{98387.5} & \textbf{96387.5} & \textbf{-147220.9} & \textbf{-122516.0} & \textbf{-8261545.0} & \textbf{-216913.3} & \textbf{-359540.6} \\ 
			& & -- & -- & -- & -- & -- & -- & -- & -- & -- & -- & 46.64 & 66.88 & 1763.44 & 1554.43 & 1800.09 & 221.58 & 241.02 & 4801.40 & 199.49 & 505.86 \\    
			\midrule	
			\multirow{9}{*}{\rotatebox{90}{\emph{TC\_ALL}}} & 
			\multirow{2}{*}{MFEA} 
			& 22900.7 & 30536.2 & 32900.5 & 31283.3 & 23500.5 & 646.3 & 725.8 & 680.3 & 1110.3 & 888.4 & 4163.0 & 6852.0 & 101420.5 & 102985.3 & 101450.7 & -135521.5 & -120100.0 & -8111898.9 & -210817.6 & -351378.5 \\ 
			& & 950.22 & 629.60 & 800.15 & 1218.84 & 794.38 & 43.07 & 30.12 & 45.62 & 34.12 & 47.01 & 115.07 & 86.04 & 3140.34 & 1135.52 & 1279.0 & 3899.10 & 1690.5 & 5639.45 & 2054.82 & 1086.5 \\  
			\cmidrule{2-22}			
			& \multirow{2}{*}{MFEA-II} 
			& 22529.6 & 29250.4 & 31994.2 & 28293.1 & 22121.0  & 615.8 & 700.8 & 676.8 & 1030.7 & 895.6 & 4048.6 & 6730.0 & 98389.0 & 101069.0 & 99064.0 & -142411.0 & -121054.2 & -8158697.3 & -213935.5 & -353615.6 \\  
			& & 830.62 & 374.02 & 700.10 & 638.58 & 312.59 & 33.99 & 36.64 & 33.94 & 18.13 & 37.60 & 79.19 & 114.19 & 2114.12 & 1977.07 & 1888.72 & 1888.72 & 803.16 & 3997.07 & 1237.81 & 2401.57 \\  
			\cmidrule{2-22}
			& \multirow{2}{*}{MFCGA} 
			& 22115.5 & 28610.3 & 32595.8 & 27899.7 & 21705.5 & 594.4 & 667.6 & 619.7 & 1002.9 & 806.0 & 3989.0 & 6587.4 & 98791.5 & 99184.5 & 99176.5 & -145297.3 & -121611.6 & -8206310.4 & -215327.8 & -354694.3 \\ 
			& & 272.42 & 277.61 & 443.61 & 316.56 & 245.33 & 9.76 & 11.87 & 13.36 & 14.10 & 14.73 & 36.25 & 57.60 & 1725.35 & 1073.10 & 1789.26 & 582.47 & 446.60 & 7915.96 & 571.35 & 1122.70 \\
			\cmidrule{2-22}
			& \multirow{2}{*}{AT-MFCGA} 
			& \textbf{21637.3} & \textbf{27769.3} & \textbf{30998.3} & \textbf{27309.2} & \textbf{21310.1} & \textbf{588.6} & \textbf{665.6} & \textbf{608.2} & \textbf{1001.9} & \textbf{796.6} & \textbf{3970.2} & \textbf{6585.4} & \textbf{97814.0} & \textbf{98910.5} & \textbf{97743.0} & \textbf{-147310.9} & \textbf{-122518.2} & \textbf{-8260978.8} & \textbf{-217034.8} & \textbf{-359566.4} \\ 
			& & 242.37 & 377.63 & 385.3 & 435.68 & 303.67 & 9.51 & 12.65 & 10.93 & 19.96 & 13.70 & 42.51 & 66.94 & 1419.86 & 1391.85 & 1771.85 & 103.11 & 245.84 & 4121.90 & 212.59 & 530.42 \\    
			\midrule	            
			& Optima & 21282 & 26524 & 29368 & 26524 & 20749 & 554 & 629 & 568 & 945 & 744 & 3744 & 6124 & 88900 & 91420 & 88700 & -163219 & -140678 & -9182291 & -254568 & -413948\\ \bottomrule
		\end{tabular}
	}
	\label{tab:results}
\end{table*}

Several interesting findings emerge after a thorough inspection of the outcomes in Table \ref{tab:results}. First of all, we clearly observe that AT-MFCGA is, overall, the best performing method. The performance shown by its non-adaptive version (MFCGA) is also worth noting, as it outperforms both MFEA and MFEA-II in most instances. Finally, the method that performs the worst is MFEA. Looking more closely at the results, the test case $TC\_ALL$ seems particularly interesting, as in it AT-MFCGA attains the best results in all its compounding 20 instances. This trend also holds in other scenarios. Finally,  and although it is not the main goal of this experimentation, it is also interesting to remark that the difference between the known optima and the average results obtained by AT-MFCGA is 1.60\% for the best case (\texttt{KroA100}), and 10,02\% for the worst instance (\texttt{Kra30a}). This is quite remarkable considering the technique is devised to simultaneously optimize 20 tasks with just 300 individuals.

Moreover, if we analyze the results reached for all test environments comprising 5 and 10 tasks and in \emph{TC\_ALL}, MFEA-II and AT-MFCGA seem to scale better and are more resilient to modifications in the problem instances to solve. Specifically, the results of MFCGA degrade significantly in some cases, obtaining the worst outcome in the last test case. This is more noticeable in instances of higher dimensions, such as \texttt{KroA150} or \texttt{KroA200}. The same degradation can be observed in \texttt{Kra30a}, \texttt{Kra30b} and \texttt{Kra32}. This phenomenon does not occur in the case of AT-MFCGA, which maintains its performance in every multitasking environment, even improving it in some cases for \emph{TC\_ALL}. This can be seen in tasks such as \texttt{KroA100} or \texttt{N-t59d11xx}, in which AT-MFCGA attains the best outcomes in last test case. As mentioned, a similar behavior can be also observed in the case of MFEA-II, which outperforms MFCGA in some specific tasks of \emph{TC\_ALL}, such as \texttt{KroA200}, \texttt{Kra30a} and \texttt{Kra32}, something not characteristic for the rest of the environments. This situation is symptomatic of the adaptability of both AT-MFCGA and MFEA-II and evinces the superiority of these adaptive methods when compared to their static versions MFCGA and MFEA.

With the aim of verifying the statistical relevance of the reported performance gaps, we follow the guidelines in \cite{latorre2020fairness} and perform two different tests with the outcomes obtained for the last test case. We have chosen this last environment since we have considered it to be the most representative and demanding one within our experimental setup. Results of both tests can be found in Table \ref{tab:resultsStats}. To begin with, the Friedman's non-parametric test for multiple comparisons allows us to prove whether differences among the results obtained by all reported methods can be declared to be statistically significant or not. The first column of Table \ref{tab:resultsStats} shows the mean ranking returned by this non-parametric test for each of the compared algorithms (the lower the rank, the better the performance). The outcomes of this test support our above statements: AT-MFCGA is the best performing solver. Additionally, the obtained Friedman statistic is $55.62$. Taking into account that the confidence interval is set to 99\%, the critical point in a $\chi^2$ distribution with 3 degrees of freedom is $11.34$. Thus, since $55.62>11.34$, it can be concluded that the differences among the results reported by the four compared algorithms are statistically significant, with AT-MFCGA ranking the lowest (the best). Besides that, in order to evaluate the statistical significance of the better performance of AT-MFCGA, the Holm's post-hoc test has been conducted and used as control algorithm. The unadjusted and adjusted $p$-values obtained as a result of the application of this procedure are depicted in the second and third columns of Table \ref{tab:resultsStats}. Analyzing this information, and taking into account that all the $p$-values are lower than $0.01$, it can be concluded that AT-MFCGA is significantly better at a 99\% confidence level.

In addition to the quantitative analysis of the results provided in this section, we will now introduce a detailed examination of the genetic transferability detected by the AT-MFCGA among studied test cases.
\begin{table}[t]
	\centering
	\caption{Results of the Friedman's non-parametric tests, and unadjusted and adjusted $p$-values obtained through the application of Holm's post-hoc procedure using AT-MFCGA as control algorithm.}
	\vspace{3mm}
	\label{tab:resultsStats}
	\scalebox{0.8}{
		\begin{tabular}{cccc}
			\toprule
			& Friedman's Test & \multicolumn{2}{c}{Holm's Post Hoc}\\
			\midrule
			Algorithm & Rank & Unadjusted $p$ & Adjusted $p$\\
			\midrule
			MFEA & 3.95 &0&0\\
			MFEA-II & 2.90 &0.000003 &0.000007\\
			MFCGA & 2.15 & 0.0004849 & 0.004849\\
			AT-MFCGA& 1.00 & -- & --\\
			\bottomrule
		\end{tabular}
	}

\end{table}

\subsection{Analysis of the Genetic Transfer between Tasks}\label{sec:exp_gen}

In this section we analyze the genetic transfer across the different instances considered in this experimentation. To conduct this examination, we focus our attention on the proposed AT-MFCGA. Furthermore, we perform four separate analyses, one for each of the problems considered. For this reason, we will use the inter-task interactions occurred along the 20 repetitions of the four tests cases dedicated to each problem: \texttt{TC\_TSP}, \texttt{TC\_VRP}, \texttt{TC\_QAP} and \texttt{TC\_LOP}. We have chosen these test cases in order to concentrate on positive and negative transfers within instances belonging to the same family of combinatorial optimization problems. As mentioned at the beginning of Section \ref{sec:exp}, as we are dealing with different problems, we assume the existence of negative transfer in every inter-problem interaction. Thus, the main goal of this study is i) to get a glimpse of the positive knowledge transfer among problem tasks; ii) to discover synergies between them; and iii) to empirically gauge inter-task interactions.

It is interesting to mention here that both MFCGA and AT-MFCGA are particularly well suited for conducting this genetic transfer analysis. This is so by virtue of the replacement strategy of these methods, namely, the \textit{local improvement selection mechanism} (Section \ref{sec:MFCGAprel}). As such, in MFCGA and AT-MFCGA an individual $\mathbf{x}_i$ of the population is only replaced if any of the solutions created as a result of \texttt{crossover} ($\mathbf{x}_i^{crossover}$) or \texttt{mutation} ($\mathbf{x}_i^{mutation}$) outperforms $\mathbf{x}_i$ in terms of its best performing task (skill factor). As such, in cases in which $\mathbf{x}_i^{crossover}$ replaces $\mathbf{x}_i$ we can ensure that a positive transfer of genetic material has occurred from $\mathbf{x}_j$ to $\mathbf{x}_i$ (see Section \ref{sec:MFCGA} and Algorithm \ref{alg:MFCGA} for details on the notation). In the context of the problems considered, this transfer is materialized through the insertion of part of $\mathbf{x}_j$ into the genetic structure of $\mathbf{x}_i$, conceiving this process as a positive contribution of task $\tau^j$ to task $\tau^i$.

\begin{figure}[h!]
	\centering
	\subfloat[]{\includegraphics[width=0.5\hsize]{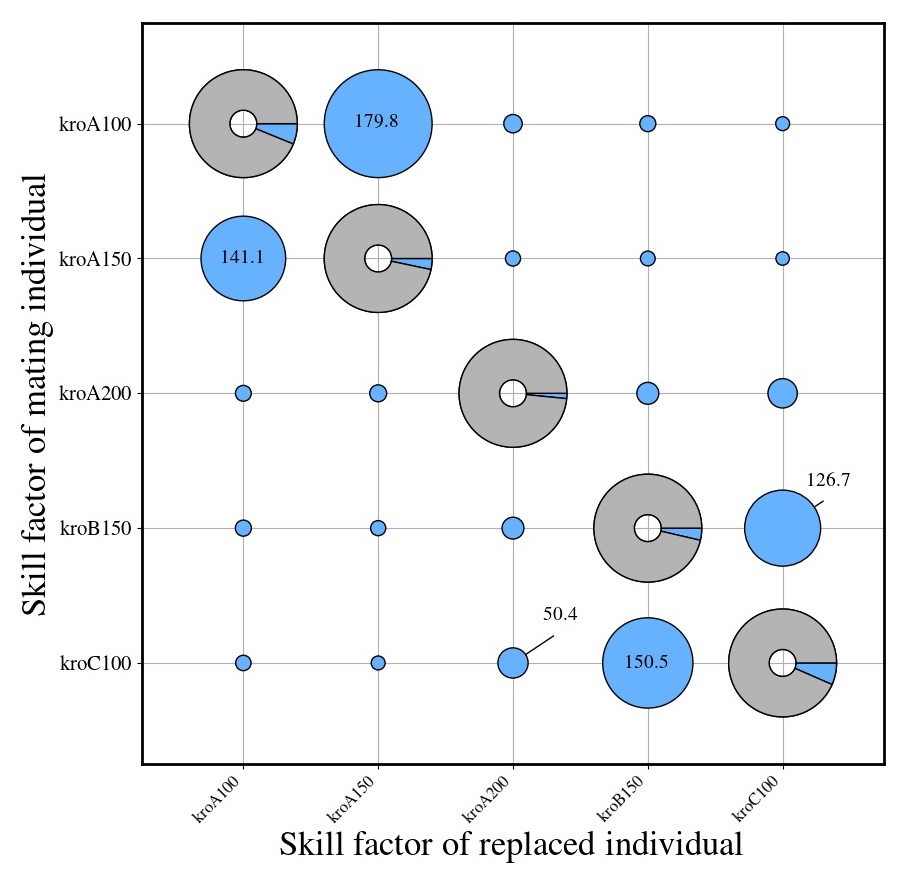}}
	\hfill
		\subfloat[]{\includegraphics[width=0.5\hsize]{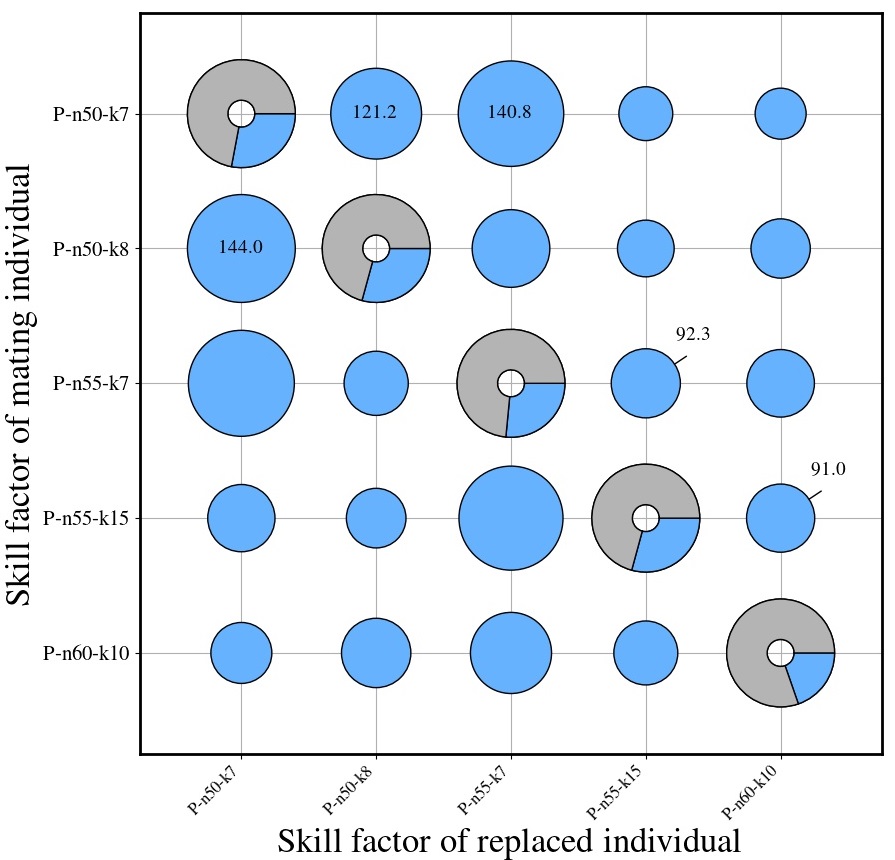}}\newline
			\subfloat[]{\includegraphics[width=0.48\hsize]{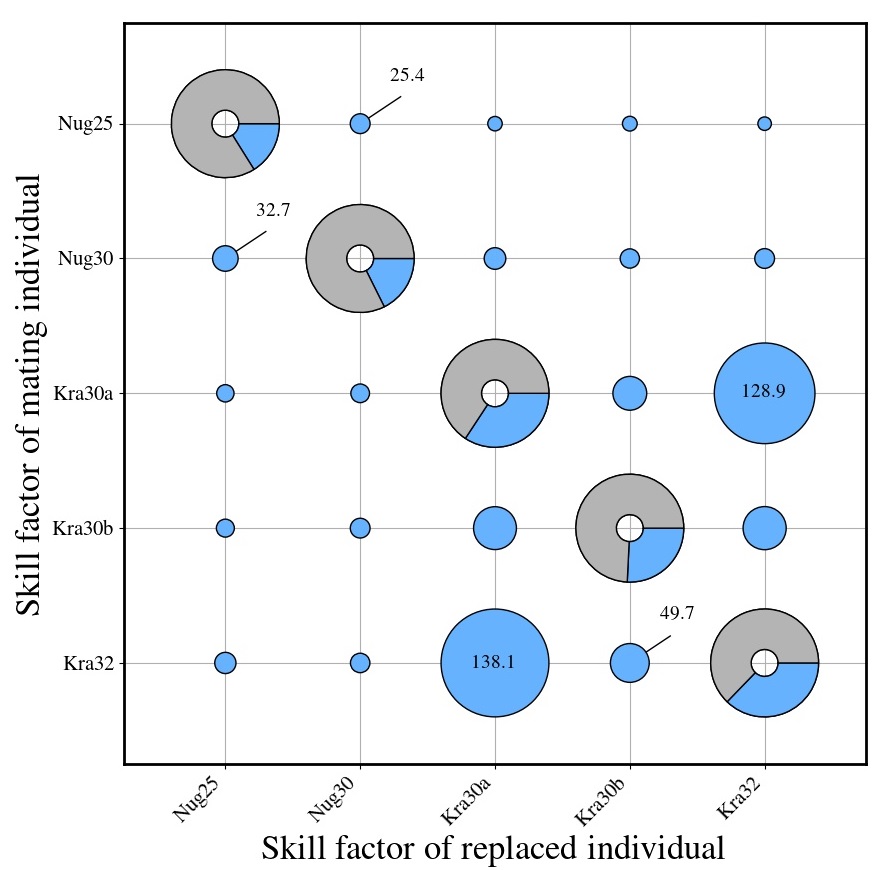}}\hfill 
				\subfloat[]{\includegraphics[width=0.51\hsize]{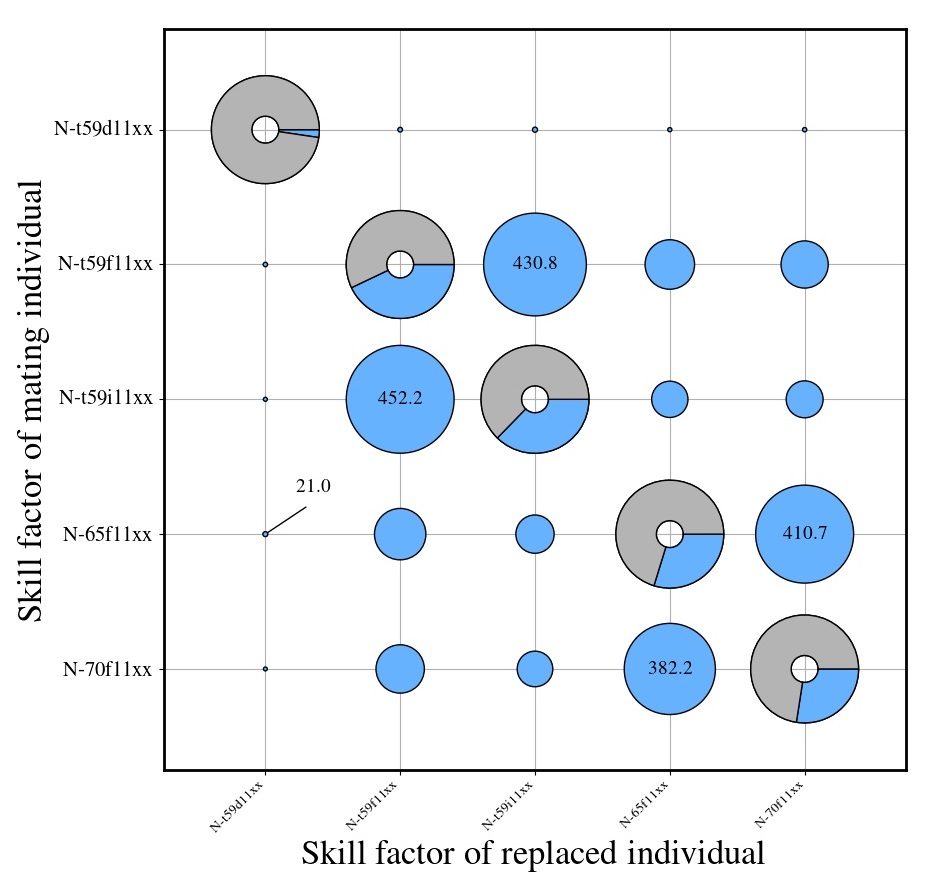}}
	\caption{Average intensities of genetic transfer between (a) TSP; (b) CVRP; (c) QAP; (d) LOP instances. Radius of each circle is proportional to the average amount of positive transfers per run in which an individual with the skill factor of the column label and an individual whose skill factor is indicated in the row label. Circles in the diagonal represent the sum of all inter-task (blue portion) and intra-task (gray portion) exchanges.}
	\label{fig:matrix_influence}
\end{figure}

Bearing in mind the above consideration, Figures \ref{fig:matrix_influence}.a to \ref{fig:matrix_influence}.c illustrate the number of positive genetic transfer episodes between each pair of tasks. In these plots, the radius of each circle is proportional to the average amount of transfers per run in which an individual with the skill factor of the column label has shared some of its genetic material with an individual whose skill factor is indicated in the row label. Furthermore, circles placed on the diagonal line symbolize the sum of all intra-task (gray portion) and inter-task (blue portion) exchanges. An intra-task transfer occurs when the genetic exchange is produced between individuals with the same skill factor.

The above figures reveal several interesting findings. The first one is the confirmation of the existence of synergies between the considered instances. An exemplifying few examples can be seen in \{\texttt{kroA100},\texttt{kroA150}\}, \{\texttt{P-n50-k7},\texttt{P-n55-k7}\}, \{\texttt{kra30a},\texttt{kra32}\} and \{\texttt{N-t59f11xx},\texttt{N-t59i11xx}\}, which evince an intense synergy over the search. We can hence confirm that the genetic transfer among these pairs of tasks (and others for which similar conclusions can be drawn) have contributed to the multitasking search process.

The second conclusion is related to the negative transfer and to those task pairs in which the intensity of genetic material exchange is almost negligible. Examples such as \{\texttt{kroA100},\texttt{kroC100}\}, \{\texttt{Nug25},\texttt{kra32}\} or any relation with \texttt{N-t59d11xx} are representative of this situation. This fact unveils that the exchanges of genetic material among these tasks can be considered to be negative \cite{bonilla2008multi}, not contributing at all to the convergence of the search process. In task pairs such as \{\texttt{Nug25},\texttt{kra32}\} or \{\texttt{N-t59d11xx},\texttt{N-t59f11xx}\}, this negative knowledge sharing could be expected beforehand, since the similarity between these two instances as per Table \ref{tab:similarity} is very low (0\% and 2\%, respectively).

Unexpectedly, the proliferation of unprofitable transfers in pairs such \{\texttt{kroA100}, \texttt{kroA200}\}, \{\texttt{kroA150},\texttt{kroA200}\} or \{\texttt{kroA200},\texttt{kroB150}\} can be contradictory with respect to the information depicted in Table \ref{tab:similarity}. The complementarity in the structure of these task-pairs can be considered as high (66\%, 57\% and 66\%, respectively). However, the inter-task interaction for these pairs depicted in Figure \ref{fig:matrix_influence}.a is practically nonexistent. This finding contradicts some influential studies \cite{gupta2015multifactorial}. Indeed, assessing the correlation in the composition of the aforementioned pairs, we can arguably confirm that the so-called \textit{partial domain overlap} exists \cite{gupta2017insights}. In other words, the domains of pairs such as \{\texttt{kroA100},\texttt{kroA200}\}, \{\texttt{kroA150},\texttt{kroA200}\} or \{\texttt{kroA200},\texttt{kroB150}\} partially overlap because of the existence of set of features which are common to both tasks.

To shed further light on this unexpected mismatch, we have conducted a deeper analysis of the 5 TSP instances. In this second study we focus our attention on the correlation between the best known solutions of such instances, for which we resort to quantitative measures proposed in recent research dedicated to continuous optimization problems \cite{da2017evolutionary,zhou2018study}. The specific definition used for \textit{intersection} or \textit{overlapping} is the one provided in \cite{da2017evolutionary}: two tasks partially overlap with each other if \textit{the global optima of the two tasks are identical in the unified search space with respect to a subset of variables, and different with respect to the remaining variables}. 

In Table \ref{tab:similarityBestSolutions} we summarize the genetic similarities found in the optimal solutions of the considered TSP instances. For the sake of completeness, we complement this table with the same data regarding the VRP, aiming at strengthening the conclusions drawn from this second analysis. We have highlighted in blue those cells corresponding to tasks that have elicited a significant positive inter-task knowledge transfer, as given in Figures \ref{fig:matrix_influence}.a and \ref{fig:matrix_influence}.b: the more intense the inter-task activity was, the more intense the blue used for coloring the entry in Table \ref{tab:similarityBestSolutions} will be. 

Several interesting trends can be observed in Table \ref{tab:similarityBestSolutions}. To begin with, pairs with the highest positive transfer activity expose a significant overlap in their optimal solutions. This statement is particularly visible in cases such as \{\texttt{kroA100},\texttt{kroA150}\}, \{\texttt{kroC100},\texttt{kroA200}\}, \{\texttt{P-n50-k7},\texttt{P-n55-k7}\} or \{\texttt{P-n50-k7},\texttt{P-n50-k8}\}. Likewise, pairs with a lower or a non existent level of overlap between their optimal solutions show less intensity on their positive genetic transfer. Arguably, TSP cases such as \{\texttt{kroA100},\texttt{kroB150}\} or \{\texttt{kroA200},\texttt{kroA150}\}, and the VRP instance pair \{\texttt{P-n55-k15},\texttt{P-n60-k10}\} are instances that buttress this fact. On the contrary, a few specific examples do not comply with this observation. This is so due to the randomness inherent in any meta-heuristic algorithm such as AT-MFCGA. 
\begin{table}[ht!]
	\centering
	\caption{Genetic complementarities among the best known solutions of the TSP instances utilized in the experimentation}	
	\resizebox{0.8\columnwidth}{!}{
		\begin{tabular}{cccccc}
			\toprule
			Instance & kroA100 & kroA150 & kroA200 &  kroB150 & kroC100\\
			\midrule
			kroA100 & \cellcolor{gray!25} & \cellcolor{cyan!60}\textbf{32\%} &  5\% & 0\% & 0\%\\
			kroA150 & \cellcolor{gray!25} & \cellcolor{gray!25} & 3\% & 0\% & 0\%\\
			kroA200 & \cellcolor{gray!25} & \cellcolor{gray!25} & \cellcolor{gray!25} & 3\% & \cellcolor{cyan!40}\textbf{21\%}\\ 
			kroB150 & \cellcolor{gray!25} & \cellcolor{gray!25} & \cellcolor{gray!25} & \cellcolor{gray!25} & \cellcolor{cyan!25}\textbf{10\%}\\
			\toprule
			\toprule
			Instance & P-n50-k7 & P-n50-k8 & P-n55-k7 & P-n55-k15 & P-n60-k10\\
			\midrule
			P-n50-k7 & \cellcolor{gray!25} & \cellcolor{cyan!60}\textbf{51\%} &  \cellcolor{cyan!60}\textbf{59\%} & \cellcolor{cyan!35}\textbf{26\%} & \cellcolor{cyan!35}\textbf{34\%}\\
			P-n50-k8 & \cellcolor{gray!25} & \cellcolor{gray!25} & \cellcolor{cyan!40}\textbf{46\%} & \cellcolor{cyan!25}\textbf{32\% }& \cellcolor{cyan!25}\textbf{30\%}\\
			P-n55-k7 & \cellcolor{gray!25} & \cellcolor{gray!25} & \cellcolor{gray!25} & \cellcolor{cyan!60}\textbf{31\%} & \cellcolor{cyan!25}\textbf{37\%} \\ 
			P-n55-k15 & \cellcolor{gray!25} & \cellcolor{gray!25} & \cellcolor{gray!25} & \cellcolor{gray!25} & \cellcolor{cyan!25}\textbf{24\%}\\
			\bottomrule
		\end{tabular}
	}
	\label{tab:similarityBestSolutions}
\end{table}

This second analysis leads to another interesting discovery: in evolutionary multitasking, positive transfers are strictly driven by the degree of intersection between the best solutions of tasks involved in the transfer. We have tested that overlapping degrees higher than 10\% suffice for guaranteeing a minimum positive push from one task to another. Moreover, instances with a greater degree of overlap are prone to showing more intense knowledge sharing. For this reason, the sole complementarity in the structure of problem instances is concluded to be irrelevant for the existence of positive genetic transfer between tasks, as has been proven empirically in our experiments.

\section{Grid Rebuilding Mechanism: Improved Knowledge Exchange and Visual Explainability of Synergies among Tasks}\label{sec:exp_rebuilding}

\begin{figure}[ht!]
	\centering
	\subfloat[]{\includegraphics[width=0.45\hsize]{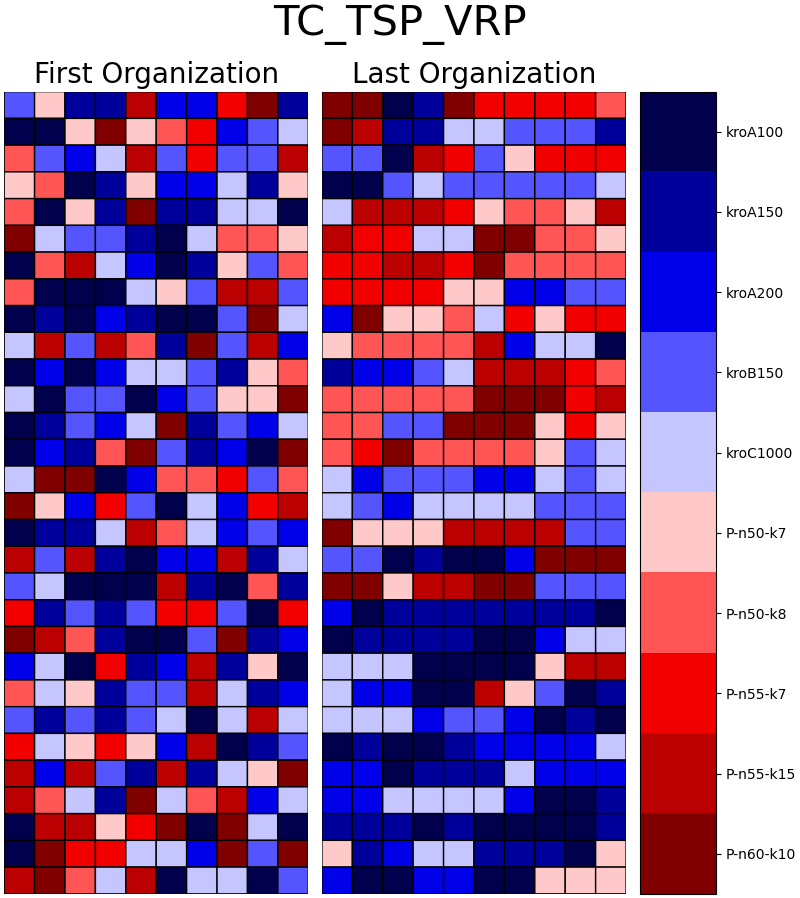}}\hfill
	\subfloat[]{\includegraphics[width=0.45\hsize]{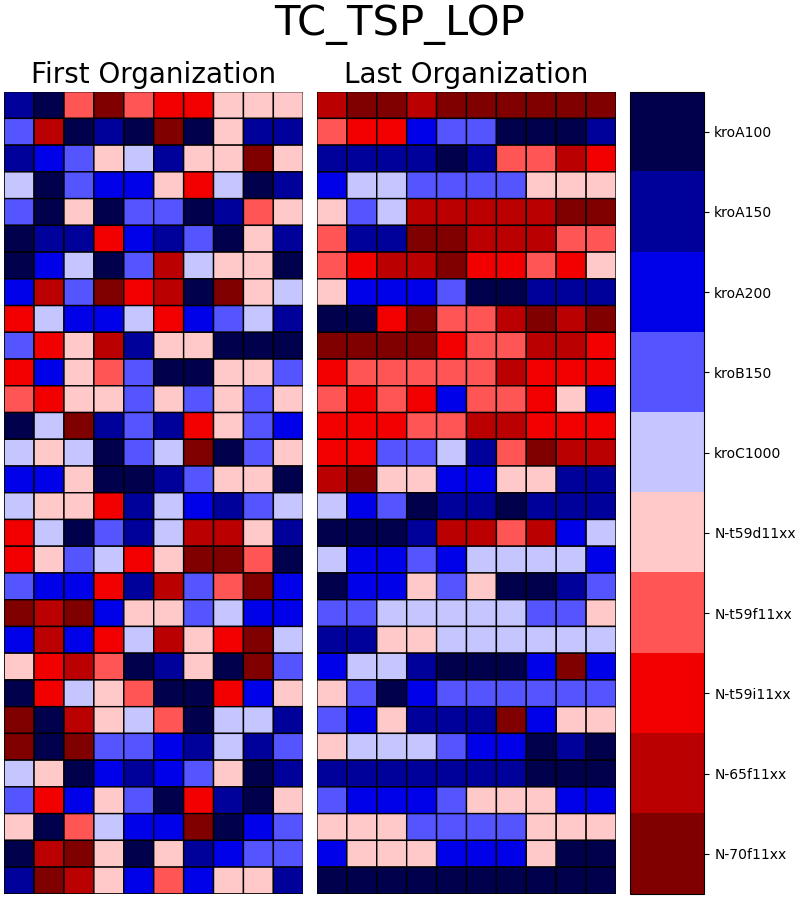}}\newline
	\subfloat[]{\includegraphics[width=0.45\hsize]{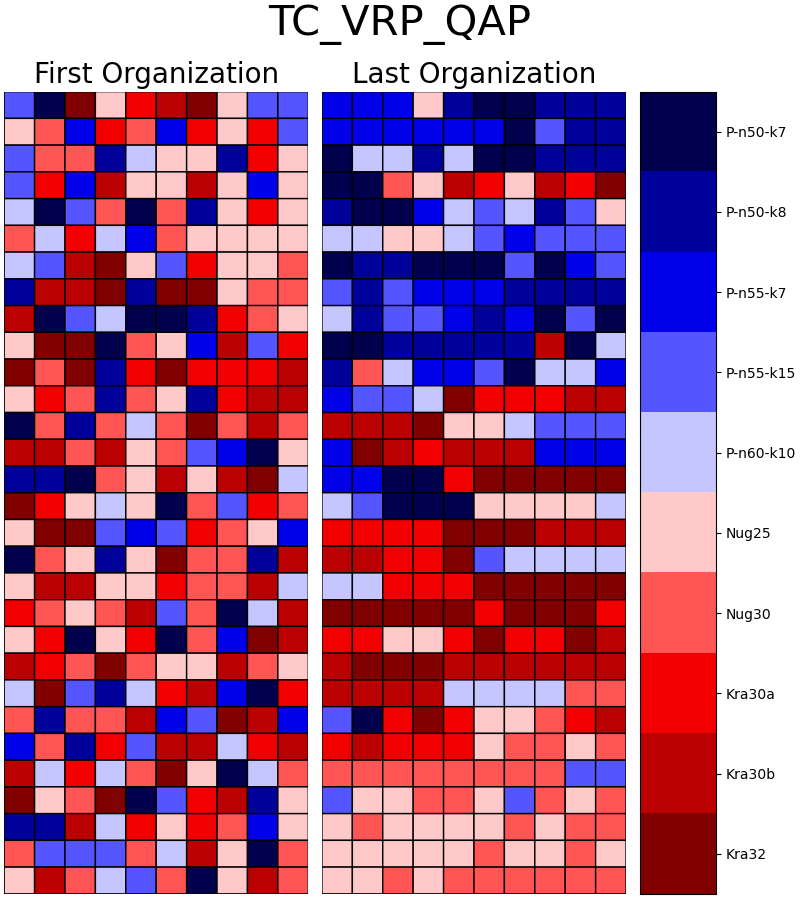}}\hfill
	\subfloat[]{\includegraphics[width=0.45\hsize]{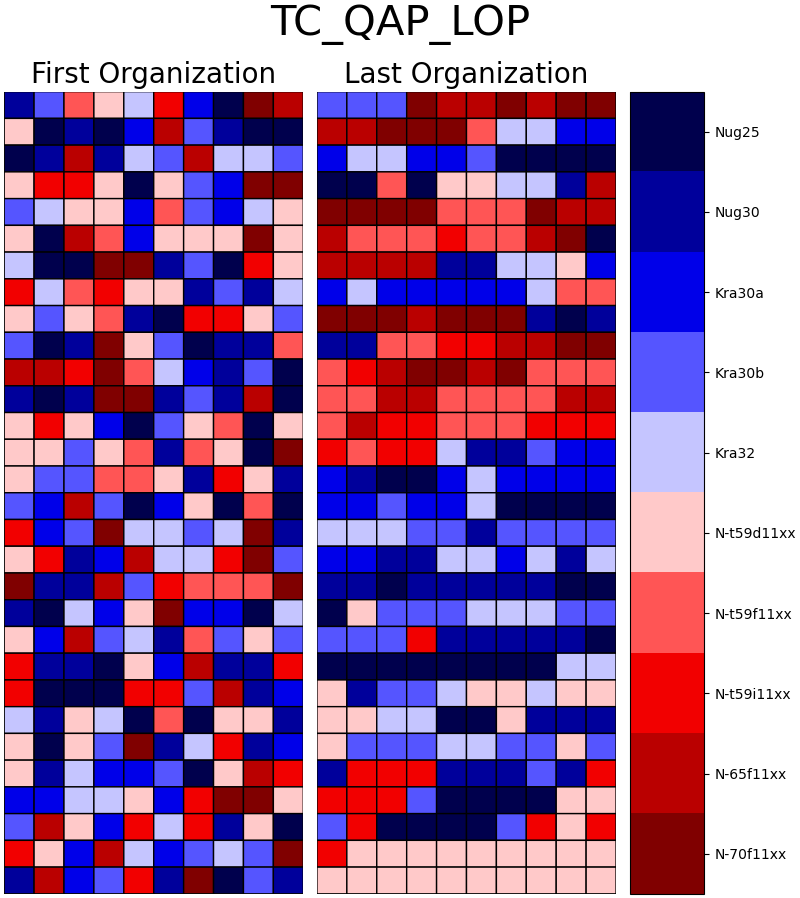}}
	\caption{First and last organizations of the cellular grid for (a) \texttt{TC\_TSP\_VRP}; (b) \texttt{TC\_TSP\_LOP}; (c) \texttt{TC\_VRP\_QAP}; (d) \texttt{TC\_QAP\_LOP}. The color of each cell depicts the skill task of the solution placed in that position. Considering that each grid is composed of $300$ individuals, the square placed in the upper-left corner represents individual $\mathbf{x}_1$, while the element in the lower-right corner represents solution $\mathbf{x}_{300}$.}
	\label{fig:grids}
\end{figure}

Finally, in a separate section we will briefly discuss the influence Grid Rebuilding has in the grid structure of AT-MFCGA. Our purpose is to demonstrate how AT-MFCGA autonomously reorganizes its whole population based on the real-time analysis of the genetic transfer produced during the execution. In this regard, one could intuitively think that AT-MFCGA reallocates the set of individuals in the cellular grid, composing new neighborhoods of interrelated tasks. However, the goal of this mechanism is not to create grids that are fully composed of individuals optimizing the same tasks. For this reasons a \textit{roulette wheel selection} procedure has been used (Section \ref{sec:AT-MFCGA} for further details). Thus, in order to avoid the premature convergence and to promote diversity in the cellular neighborhoods, our goal with this mechanism is to create grids composed mainly of synergistic tasks, but not excluding the inclusion of non-related tasks.

Figures \ref{fig:grids}.a to \ref{fig:grids}.d show the initial and final organizations of the cellular grid of AT-MFCGA in 4 of the 6 test cases, composed of two different problems:  \texttt{TC\_TSP\_VRP}, \texttt{TC\_TSP\_LOP}, \texttt{TC\_VRP\_QAP}, and \texttt{TC\_QAP\_LOP}. We have chosen these four cases in order to facilitate the visual understanding of the grid rebuilding mechanism. For example, \texttt{TC\_all} would be hard to visualize in a such figure as it contains 20 different tasks. Furthermore, in order to generate these figures, we have gathered the information resulting from the first of the $20$ runs executed in each test case. Each of the colored squares corresponds to a specific individual in the grid. Since each of the test cases represented in these figures comprises two problems, tasks referring to the first problem are filled with a red-color palette, whereas instances corresponding to the second task are colored using different tones of blue. This information is shown in every figure. Considering that each $10\times 30$ grid comprises $300$ individuals (see Table \ref{tab:Parametrization1} for more detail), the square placed in the upper-left corner corresponds to $\mathbf{x}_1$, and the element in the lower-right corner depicts $\mathbf{x}_{300}$. The color of each cell represents the skill task of the solution placed in that position. Each time the \textit{Grid Rebuilding} mechanism is executed, the placement of individuals is modified by following the principles described in Section \ref{sec:AT-MFCGA}. It is worth noting that in these figures we do not explicitly show in which position a specific individual is arranged, since the tracking of a concrete solution is of no interest in this study. Instead, the purpose of these plots is to yield a general picture of how individuals with related skill tasks are placed in close positions.

In these figures we clearly discern the main philosophy of the \textit{Grid Rebuilding} mechanism: tasks belonging to the same problem (those of the same primary color) tend to be placed in adjacent positions. This is clearly verifiable in every test case: in Figures \ref{fig:grids}.a, \ref{fig:grids}.b and \ref{fig:grids}.d, for example, a group of blue individuals can be distinguished in the bottom and central part of the grid. In Figure \ref{fig:grids}.c this group block can be identified in the upper part of the grid. The same observation can be made when focusing on the red palette: red blocks can be perceived in the upper parts of Figures \ref{fig:grids}.a, \ref{fig:grids}.b, and the bottom part of Figure \ref{fig:grids}.c. It is also interesting to analyze isolated tasks, e.g. in Figure \ref{fig:grids}.d, individuals specialized in task \texttt{N-t59d11xx} are mainly concentrated in the bottom part of the matrix. Taking Figure \ref{fig:matrix_influence}.d into account, we know that \texttt{N-t59d11xx} is not complementary with any other intra-problem tasks. Furthermore, we see that no inter-problem relationship is synergistic. For this reason, all individuals optimizing \texttt{N-t59d11xx} tend to be placed in the last part of the grid.

Given these outcomes we can conclude that \textit{Grid Rebuilding} mechanism effectively reorganizes the cellular grid of AT-MFCGA, favoring the adjacency of synergistic tasks, and isolating those that do not contribute to a better convergence of their counterparts. The information contained in the grid after the execution of this mechanism presents the relationships between tasks, grouping them together spatially. This offers an explanation of what AT-MFCGA discovers during the search, which helps understand the evolution of the knowledge grasped by the algorithm, disregarding the technical background of the user at hand.

\section{Conclusions and Future Research}\label{sec:conc}

This paper has elaborated on the design, implementation and performance analysis of an Adaptive Transfer-guided Multifactorial Cellular Genetic Algorithm (AT-MFCGA) suited to dealing with multitasking scenarios in which several optimization problems must be solved by a single search process, harnessing eventual synergies between problems. Our method incorporates several key aspects that make it a promising meta-heuristic for multitasking setups: 1) a neighborhood relationship induced on a grid arrangement of the individuals, which allows the coverage of the evolutionary crossover operator to be controlled; and 2) two adaptive mechanisms in order to efficiently face negative knowledge transfers: \textit{Grid Rebuilding} and \textit{Multi-Mutation}.

In order to quantitatively assess the performance of the proposed approach, we have designed 11 multitasking environments comprising 20 different instances of 4 combinatorial optimization problems (TSP, CVRP, QAP and LOP), over which the quality of solutions produced by AT-MFCGA has been compared to that of MFEA, MFEA-II and the non-adaptive MFCGA. The obtained results  verify that AT-MFCGA is a promising method that performs better (with statistical significance) than the other methods considered in the benchmark. Furthermore, an additional analysis of the inter-task genetic transfer produced during the search process has been carried out, and shows that the empirical crossover counts between tasks are in accordance with the estimated overlap of their optimal solutions, hence uncovering the complexity of identifying the synergies between problems beforehand. Finally, the last stage of our experimental study shows the impact of the grid rebuilding mechanisms of AT-MFCGA, clearly depicting how individuals optimizing synergistic tasks are prone to be placed close to each other with the entire cell. This last feature of AT-MFCGA also provides a friendly interface for users to better understand relationships existing between tasks.

The findings reported in this study pave the way for several future research directions. In the short term, it is our intention to assess the efficiency of AT-MFCGA using additional problem instances. Another interesting line of research to be pursued in the near future is to exploit the information contained in the cellular grid to cope with non-stationary tasks, namely, tasks that evolve over time. We believe that the neighborhood-based structure of AT-MFCGA does not only contribute to the convergence of the overall algorithm, but also serves to detect changes in tasks that reflect the synergies among problems. Finally, we plan to address other practical scenarios suited to be tackled with multitasking approaches, such as fuzzy control systems \cite{precup2019nature}, fuzzy conformable fractional differential equations \cite{arqub2020fuzzy}, second-order boundary value problems \cite{arqub2014numerical}.

\section*{Acknowledgments}

The authors thank the Basque Government for its support through the Consolidated Research Group MATHMODE (IT1294-19), as well as the ELKARTEK program (3KIA project, ref. KK-2020/00049). Authors also acknowledge the financial support from the project SCOTT: Secure Connected Trustable Things (ECSEL Joint Undertaking, ref. 737422). Francisco Herrera would like to thank the Spanish Government for its funding support (SMART-DaSCI project, TIN2017-89517-P).

\bibliographystyle{elsarticle-harv}
\bibliography{biblio}

\end{document}